\documentclass{article}
\usepackage{cite}
\usepackage{a4wide}
\usepackage{amsfonts}
\usepackage{amsmath}
\usepackage{algorithm}
\usepackage{algpseudocode}
\usepackage{caption}
\usepackage{subcaption}
\usepackage{booktabs}
\usepackage{graphicx}
\usepackage{multirow}
\usepackage{listings}
\usepackage{url}
\usepackage[colorinlistoftodos]{todonotes}
\usepackage{authblk}

\begin{document}

	\title{Video Captioning: a comparative review of where we are and which could be the route}

\author[1]{Daniela Moctezuma}
\author[1,2]{ Tania Ramírez-delReal }
\author[1,2]{Guillermo Ruiz}
\author[1]{ Othón González-Chávez}
\affil[1]{Centro de Investigación en Ciencias de Información Geoespacial AC, Circuito Tecnopolo II , Aguascalientes, 20313, Mexico}
\affil[2]{Consejo Nacional de Ciencia y Tecnología (CONACyT), Av. Insurgentes Sur 1582, Ciudad de Mexico, 03940, Mexico}
\textit{This manuscript is currently submitted to Computer Vision and Image Understanding Journal}

	\maketitle
	\begin{abstract}
		Video captioning is the process of describing the content of a sequence of images capturing its semantic relationships and meanings. Dealing with this task with a single image is arduous, not to mention how difficult it is for a video (or images sequence). The amount and relevance of the applications of video captioning are vast, mainly to deal with a significant amount of video recordings in video surveillance, or assisting people visually impaired, to mention a few.
		To analyze where the efforts of our community to solve the video captioning task are, as well as what route could be better to follow, this manuscript presents an extensive review of more than 105 papers for the period of 2016 to 2021. As a result, the most-used datasets and metrics are identified. Also, the main approaches used and the best ones. We compute a set of rankings based on several performance metrics to obtain, according to its performance, the best method with the best result on the video captioning task. Finally, some insights are concluded about which could be the next steps or opportunity areas to improve dealing with this complex task.

	\end{abstract}


\section{Introduction}

The incredible explosion of data from the Internet (images, text, videos, etc.) challenges us to deal with it efficiently and appropriately. One of the essential tasks of the research community is generating data understanding technology, for instance, to comprehend a video sequence from surveillance systems. The number of applications that could be beneficial with this technology is enormous, for instance, content retrieval systems, smart video surveillance, and computer-human interface systems, among others.

Nowadays, machine learning methods have well-defined areas, {\it e.g.}, natural language processing, and artificial vision. One of the most famous and complex tasks is image captioning, which describes the image content using natural language sentences like humans would do. We experience, as humans, that describing visual content is very simple, but for machines, it is a difficult task because there are many aspects to consider, such as the objects in the image, the relations between them, the semantic meaning, and context data, along with others.

There are tremendous advances in image captioning tasks, mainly tackled by deep learning approaches. In this sense, the video captioning task emerged with little attention but showed exciting results with the advent of deep learning algorithms. Besides the complexity of describing a single image's content, describing a video, or images sequences is more challenging because the time variable becomes crucial to determine the relationship between objects, detecting the actions, and so on.

Nowadays, several approaches are dealing with this task, for instance those exploiting temporal cues~\cite{shi2019watch}, \cite{mun2019streamlined},\cite{guo2019exploiting}, motion~\cite{li2018jointly, wang2018m3}, \cite{long2018video}, action recognition~\cite{ramanishka2016multimodal}, \cite{shetty2016frame}, \cite{hu2019hierarchical}, people trajectories~\cite{qi2019sports}, \cite{zhang2019object}, events detection~\cite{krishna2017dense}, \cite{mun2019streamlined}, \cite{zhang2019show},  optical flow~\cite{chen2019motion}, \cite{zhou2018end}, audio~\cite{xu2017learning}, \cite{chen2017video}, \cite{ lee2019deep}, speech recognition~\cite{iashin2020multi}, to name a few.
As it will show in Section~\ref{sec:aboutapproaches} the most fundamental approaches are based on several deep learning architectures, and most of the works are designed as an encoder-decoder model for visual and text data.
Another essential aspect presented in many of the analyzed approaches is the well-known attention mechanism that was designed to improve the performance of the encoder-decoder model, specifically on the machine translation task.

A difference between image and video captioning is that in the video, there is a dependency between images to understand the meaning of its content. The whole sequence (or at least a subset of it) must be processed to generate the sentences describing it, on contrary in image captioning, there is no such dependence. Regardless of that, both tasks are currently challenging in machine learning and the natural language processing research community.
A possible improvement in video captioning methods' performance could be enhancing the caption generation. In~\cite{shi2020video}, the authors mention that there is a considerable advance in image's coding, but lower performance in the quality of the caption generation process.


In this review, we attempt to show the current state of the video captioning task; the primary datasets used to measure the performance of the proposed approaches, the performance metrics employed, the best-published results, and a deep discussion and analysis of them. To do this, we searched, analyzed, organized, and compared several recent papers and their reported results more comprehensively. This review was done analyzing papers from 2016 to 2021, and most of them were published in conferences or journals. Nevertheless, a few (very few) were considered even though they were published on the arXiv platform\footnote{https://arxiv.org/} to avoid dismissing nothing of the recent work.
Through this analysis, some conclusions have also been settled, highlighted some possible improvements, and finally, a complete overall comparison between the analyzed works to find the best solution for the studied period, is presented.

The manuscript is organized as follows, Section~\ref{definition} explains in formal terms, what is the video captioning task and which are its parts or components. Section~\ref{sec:performanceevaluation} incorporates all the elements involved in evaluating the proposed methods or approaches, that means the datasets and the performance metrics. The analysis and discussion are done in Section~\ref{sec:resultsandiscussion}, some aspects to be improved are described in Section~\ref{sec:areastobeimproved} as well as some possible applications. Finally, Section~\ref{sec:conclusions} gives some conclusions derived from our literature revision and comparison. In Appendix~\ref{sec:appendix} we present tables with all our raw reviewed data.

\section{Problem definition}
\label{definition}

The main goal of video captioning is to enable computers to understand what is happening on a video and build a solid relationship between that content and its corresponding natural language description~\cite{yan2019stat}. Video captioning could be seen as the automated task of generating a collection of natural language sentences that describe or explain the video's content~\cite{islam2021exploring}.

As any machine learning problem, the video captioning task can be formulated as follows. Given a pair of $(xs,ys)$ from dataset $\mathcal{D}$,  where $xs$ is a set or sequence of images $xs ={xs_1, xs_2,...,xs_n}$, and $ys$ is a sequence of words, $ys = (w_1, w_2,...,w_m)$ that describes the content of $xs$, find a model that maximizes $p(w_1, w_2,...,w_m |  xs_1, sx_2,...,xs_n)$, that is, find the highest probability of descriptive power $ys$ according with the respective sequence of images $xs$ for all pairs $(xs,ys)$ in the dataset $\mathcal{D}$.

\begin{figure*}[h]
	\centering
	{\includegraphics[width=0.75\textwidth]{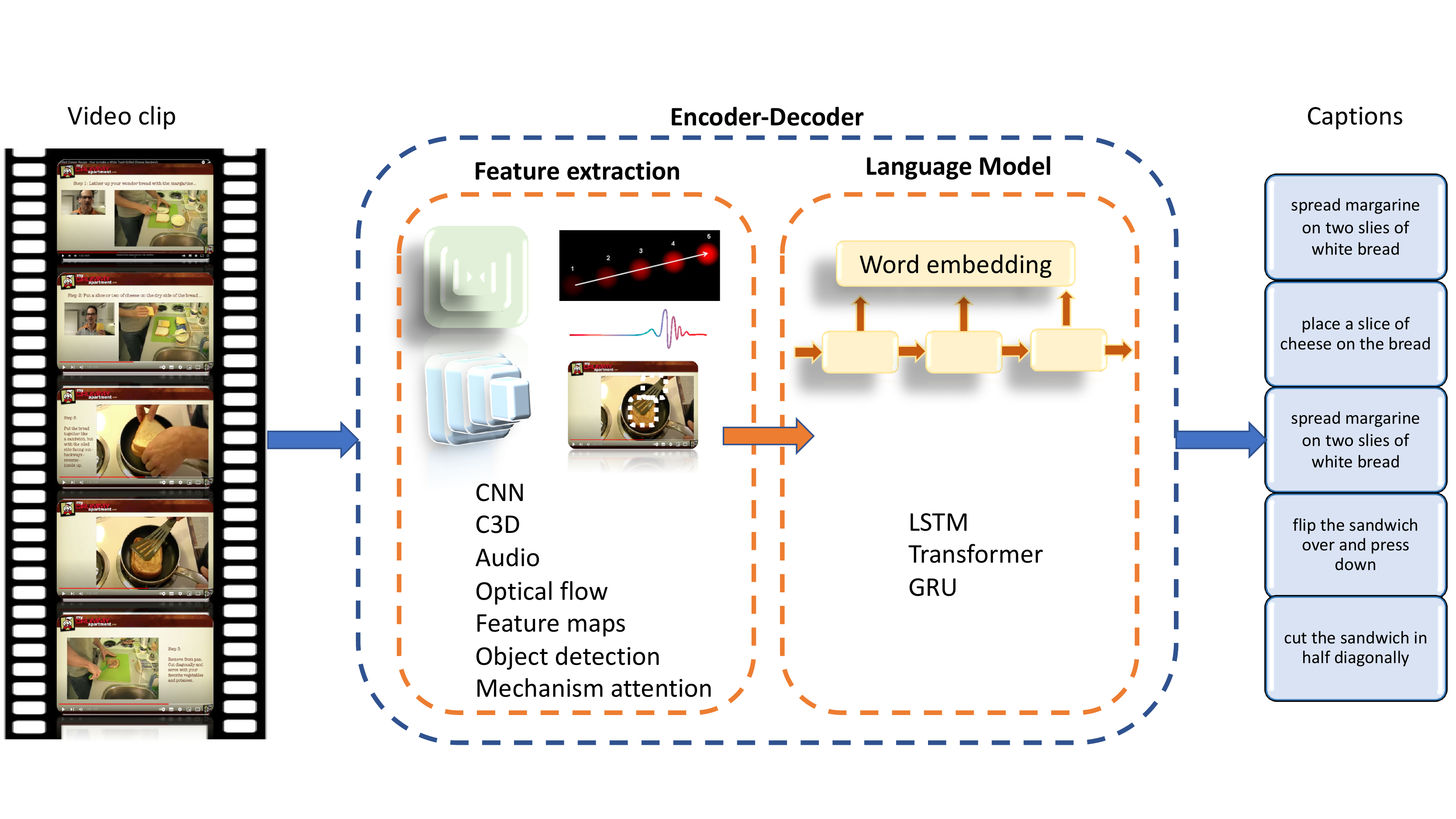}}
	\vspace{-0.2cm}
	\centering \caption{General overview of a typical solutions for video captioning task}
	\label{fig:video}
\end{figure*}

Figure \ref{fig:video} shows a general overview of the typical solutions for the video captioning task. Most of the works analyzed in this review, have an encoder-decoder framework.
The encoder extracts the video features; then, the decoder translates these attributes into the text to generate the descriptions. Different techniques are used to obtain the features. One of them is convolutional neural networks (CNN) in two and three dimensions (C3D); also, an attention mechanism is applied for temporal and spatial characteristics. Furthermore, other features are considered, such as audio, optical flow, maps, object detection, etc. The decoder does the translation applying recurrent neural networks (RNN), specifically, the variant long short-term memory (LSTM). Transformers are another architecture used, as well as the gated recurrent units (GRUs).
The video captioning models continuously improve to perform better in the training process until the generated captions become similar to the annotated ones. The attention mechanism allows supplementary data to prepare the model to predict the sentences.

A recent review published by~\cite{islam2021exploring} organizes the video captioning methods into three classes, traditional methods, machine learning, and deep learning methods.
Furthermore, the authors describe parts of the solution of the video captioning task, such as object and action recognition, objects trajectories, etc.; in our review, we will only focus on the overall video captioning task. Another review presented by~\cite{kumar2020review} described many video captioning problems, particularly in extracting multimodal features, vanishing gradients, and the training time and high resources with deep learning techniques. Also, Kumar et al. explained the possible solutions found in the literature; the papers reported employing more information related to visual, movement, audio, and semantic data contained in the video. Finally, they presented the techniques to extract the features for each data type, where deep learning is widely used, mainly VGG, Resnet, and Inception architectures. Furthermore, other methods help to save temporal information and obtain a feature map for an image sequence; likewise, they concluded that data such as the audio and the known category in videos improve the task of video captioning. A very descriptive work on the several components of proposed approaches is presented by Jain et al. in~\cite{jain2022video}.

In this paper, we tried to detect the best current solution for the video captioning problem, and more than 105 papers were analyzed in a period from 2016 to 2021. Several aspects were observed, the number of datasets used, the main metrics employed, along with the reported results. We determined which strategy could be considered the best-published solution in this topic in the period considered with several ranking approaches.

\section{Performance evaluation}
\label{sec:performanceevaluation}

As with any computational task in artificial intelligence, the performance evaluation of the different approaches is crucial to know all the solutions' strengths and weaknesses. Some aspects, such as datasets, metrics, and general conditions, are necessary for an adequate assessment.
The following section briefly describes all the datasets, from most to least used in the papers reviewed.

\subsection{Datasets}
\label{sec:datasets}
The datasets for video captioning are varied, and the majority of them are publicly available; they mainly belong to cooking or movie clips. This subsection is highlighted through the description and current dataset status regarding its availability and organization of their annotation files.

\textbf{MSVD (MS-Video YoutubeClips)}.
MSVD \cite{chen2011collecting} or YoutubeClips is one of the first datasets, proposed in the year 2011 and generated for the video captioning task. The dataset acquisition was made with Amazon's Mechanical Turk assistance.
It contains 1,970 clips and 70,028 sentences, each with 8.7 words on average; the total video duration is 5.3 hours, and it has 13,010 different terms. Although this dataset was able to be downloaded in the past, at this time its link is disabled on the official Microsoft site.

\textbf{TRECvideo Data (TRECViD)}. TRECViD \cite{khan2012natural} is conformed by videos about news, meeting, crowd, grouping, traffic, music, sports, animals, and humans interacting with objects. The dataset supplies 140 videos, and there are 20 segments for each. The description for clips has four to six sentences; other information for annotations is keywords and title. The dataset is available on the official site\footnote{trecvid.nist.gov}, and the clips and annotations are updated annually for different tasks.

\textbf{TACoS-MultiLevel}~\cite{rohrbach2014coherent}. It is based on TACoS (Saarbrücken Corpus of Textually Annotated Cooking Scenes)~\cite{regneri2013grounding}.
The dataset consists of 185 videos and 52,478 sentences in total.
The videos' content is about cooking different dishes, and their descriptions are conformed by 3 to 5 sentences.
The download is through the official site \footnote{www.mpi-inf.mpg.de/departments/computer-vision-and-machine-learning/research/vision-and-language/tacos-multi-level-corpus}, and it is essential to comment that the videos are associated with MPII Cooking 2 Dataset \cite{rohrbach15ijcv}.

\textbf{Youtube2text}. Youtube2Text \cite{guadarrama2013youtube2text} is an emerging subset of MSVD; the main difference is the specific use of the English language. Also, Guadarrama et al. proposed a unique split in adjacent videos for training and testing, 1300 and 670, respectively.
Contrary to the MSVD dataset, the derived English corpus can be found on its website\footnote{www.cs.utexas.edu/users/ml/clamp/videoDescription}.

\textbf{MPII-Movie Description corpus (MPII)}. MPII \cite{rohrbach2015dataset} is a collection of videos belonging to movies. 
The primary purpose of this dataset is to provide audio descriptions for movies.
It is composed of 94 videos, with 68337 clips and 68375 sentences, and the duration is 73.6 hours.
By filling out a request form, videos and annotations are available under demand.

\textbf{Montreal Video Annotation Dataset (M-VAD)}. M-VAD \cite{torabi2015using} incorporates 84.6 hours of 92 DVDs with 48,986 clips and 55,904 describing sentences. Its data was collected by Descriptive Video Service (DVS) encoded in DVDs. By filling out a submission format, videoclips and annotations are available under request.

\textbf{MSRVideo to Text (MSR-VTT)}. MSR-VTT \cite{xu2016msr} is one of the most extensive datasets for the video captioning task. MSR-VTT supplies 10K web video clips and 200K sentences in total; therefore, 20 different sentences describe each clip; and the length of each one is 10 to 30 seconds; the total duration is about 41.2 hours. The video content is mainly related to gaming, sports, and movies.
The official page of this dataset does not have a link that allows its download.

\textbf{Large Scale Movie Description Challenge (LSMDC)}. LSMDC \cite{rohrbach2017movie} is a subset of videos from M-VAD and MPII; also, it includes more movies.
The dataset is split to balance the movie genres; the purpose is diversity in the vocabulary.
LSMDC 2016 contains 101,046 and 7,408 training and validation clips, respectively.
The 2021 version is currently available \footnote{https://sites.google.com/site/describingmovies/}, which must be requested by sending a form to access it.

\textbf{Charades~\cite{sigurdsson2016hollywood}}. To build this dataset,
the Hollywood in Homes process, and Amazon Mechanical Turk tools were used. It can also be used for action classification and localization tasks.
The dataset contains 9,848 videos with 30.1 seconds on average length and 27,847 textual descriptions. It is available through its official site\footnote{allenai.org/data/charades}.

\textbf{ActitivityNet Captions} \cite{krishna2017dense}. It incorporates 20k videos, and the duration is 849 hours with 100k full descriptions, is based on the ActivityNet~\cite{caba2015activitynet} dataset for human activity understanding. Each clip has a duration of 10 minutes, and it contains 3.65 phrases on average. The descriptions belong to a specific time. This dataset is available for download on its official site \footnote{cs.stanford.edu/people/ranjaykrishna/densevid/}.

\textbf{Visual Attentive Script (VAS)}. VAS \cite{yu2017supervising} contains 144 video clips of 15 seconds, and each video has three sentences.
This dataset is not available to download.

\textbf{YouCookII} \cite{zhou2018towards} is a dataset with two annotations per video segment, the first description explains a recipe, and the second is a verification.
The dataset contains 2,000 clips describing 89 recipes, 13,829 sentences, and the duration is 176 hours in total; each video has 3-16 segments.
The dataset and all its information are available on its official site~\footnote{youcook2.eecs.umich.edu}.

\textbf{Fine-grained Sports Narrative (FSN)}. FSN \cite{yu2018fine} contains 2,000 videos from Youtube with an average of 3.16 sentences for each, particularly basketball playing in the NBA. The annotations consist of timestamps and a descriptive paragraph. The dataset is not available for download.

\textbf{VATEX} \cite{wang2019vatex} contains 41,250  videos and 825,000 sentences. The videos consist of human activity, and it is in two languages, English and Chinese. The data collection is from Kinetics-600, and it is combined with YouTube videos associated with human activities, the duration of each clip is about ten seconds. Amazon Mechanical Turkers made the captions according to the videos, 10 for each language. The dataset is available through its official site \footnote{eric-xw.github.io/vatex-website}.

\textbf{Sports Video Captioning Dataset-Volleyball (SVCDV)}. SVCDV \cite{qi2019sports} is another dataset related to sports, specifically volleyball. This dataset contains 55 videos from Youtube composed of 4,830 clips, with an average of 9.2 sentences for each, which means 44,436 phrases in total. The dataset is not available for download.

\textbf{Object-oriented captions}. Object-oriented captions \cite{zhu2020understanding} dataset is based on ActivityNet Captions, explicitly a subset related to the action where a game is played. The authors proposed making a new annotation to have direct relations expressed in the sentences between the objects. The dataset contains 75 videos and 534 sentences; furthermore, the descriptions are more extensive because it includes more words, verbs, and adjectives than MSR-VTT, MSVD, ActivityNet Captions, and FSN datasets. The dataset is not available for download.

\textbf{LIRIS human activities dataset}. LIRIS \cite{wolf2014evaluation} is a dataset specialized in human activity, particularly surveillance and office conditions. The original dataset contains 828 actions without captions, and it is organized into two subsets named D1 and D2. In recent work \cite{inacio2021osvidcap}, subset D2 is refreshed with descriptions of the scene action, and it supplies 167 videos with 367 annotations related to spatial, temporal, and description content. This dataset has an official site\footnote{projet.liris.cnrs.fr/voir/activities-dataset}, and it is possible to download the clips; the caption sentences are not available in any repository.

\textbf{UET Video Surveillance (UETVS)}. UETVS \cite{dilawari2021natural} is a collected database for smart surveillance purposes, it contains 1200 videos with  3-6 sentences for each video, and the descriptions were acquired from professional English writers. The dataset is not available for download.

\textbf{AGRIINTRUSION} \cite{dilawari2021natural}. This dataset contains 100 videos from YouTube, and the context is the agricultural environment. The dataset is not available for download.

All the aforementioned datasets were used to assess the proposed methods observed in our literature revision. Nevertheless, there are more and less used datasets, the most used detected from 2016 to 2021 years, are MSVD and MSR-VTT, maybe due to the diversity and variety in videos and captions. It is essential to highlight that the published articles use the terms MSVD, Youtube2Text, and YoutubeClips to refer to the same dataset; however, Youtube2Text is a subset that only contains the English language; in addition, it is important to point out the fact that the authors proposed a specific split for the evaluation of the models.
Another widely used dataset is ActivityNet Captions, mainly due to annotations in a specific time. Charades has also begun to take off to evaluate proposals and will probably have more use since it is of recent creation.

\begin{figure*}[!h]
	\centering
	\begin{tabular}{cc}
		\includegraphics[width=75mm]{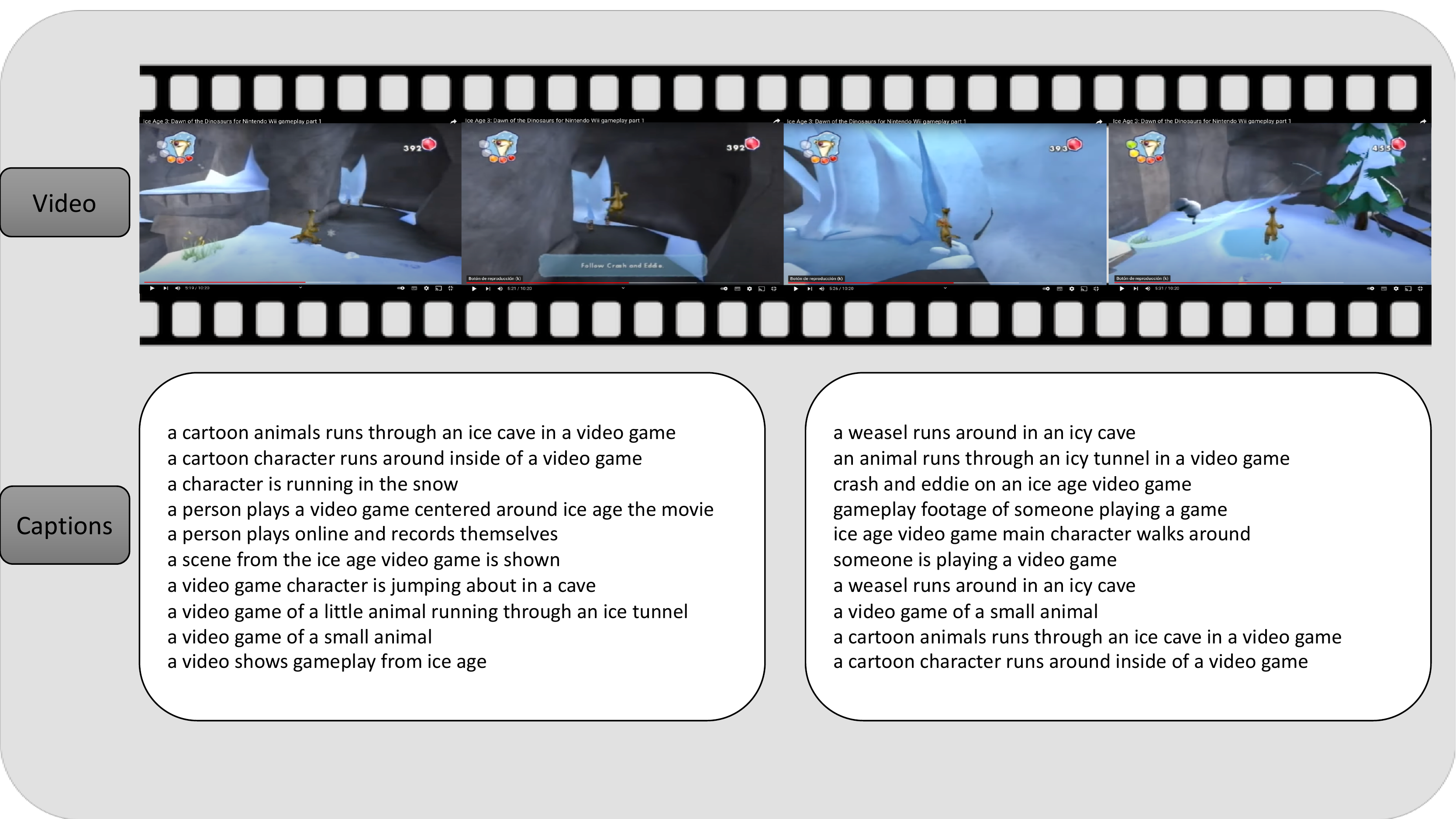} &   \includegraphics[width=75mm]{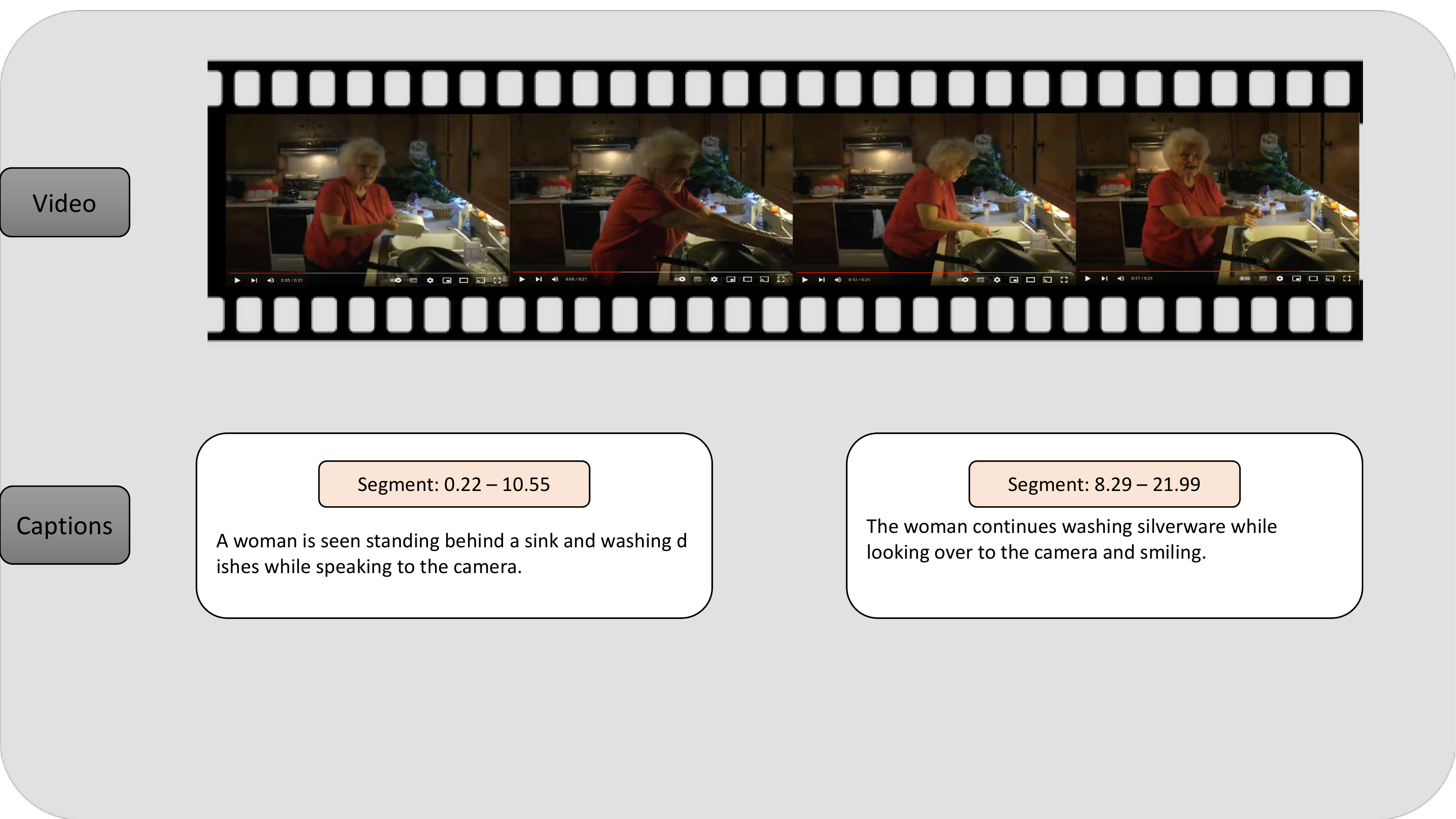} \\
		(a) Sample of MSR-VTT  & (b) Sample of ActivityNet Captions  \\[6pt]
	\end{tabular}
	\caption{Two representative samples from benchmarks for video captioning datasets with ground-truth captions}
	\label{fig:example_ds}
\end{figure*}

Figure \ref{fig:example_ds} shows a representative sample for two datasets; Figure \ref{fig:example_ds} (a) is for the MSR-VTT dataset; the clip contains 20 different sentences describing a segment of a video; Figure \ref{fig:example_ds} (b) belongs to a sample of ActivityNet Captions, which is a dataset used in dense-captioning, the clips can have distinct phrases to describe actions in overlapped segments.

In summarizing, Table \ref{tab:datasets} shows essential information about the datasets, especially the year of the creation, the videos and clips they contain, and finally, the total number of sentences. It is necessary to mention that the UETVS and AGRIINTRUSION datasets are not in the table because the total number of sentences is not available, and they are not published in a repository or website. The bolded text datasets correspond to those most used, considering a threshold of more than seven citations. The complete analysis presented in this work is focused on these seven most-used datasets.

\begin{table}[h]
	\caption{Datasets for video captioning.  $^{*}$ Available online,$^{**}$ Available through request.}
	\label{tab:datasets}
	\centering
	\resizebox{9cm}{!}
	{
		\begin{tabular}{|c|c|c|c|c|}
			\hline
			Dataset   & Year & Videos & Clips & Sentences   \\ \hline
			\textbf{MSVD} \cite{chen2011collecting} & \textbf{2011} & \textbf{1,970} & \textbf{1,970} & \textbf{70,028} \\
			TRECViD$^{*}$ \cite{khan2012natural} & 2012 & 140 & 140 &1,820   \\
			\textbf{Youtube2Text}$^{*}$  \cite{guadarrama2013youtube2text} & \textbf{2013} & \textbf{1,970} & \textbf{1,970} & \textbf{70,028} \\
			TACoS-MultiLevel$^{*}$ \cite{rohrbach2014coherent}& 2014 & 185 & 16,145 & 52,478 \\
			\textbf{MPII-MD}$^{**}$ \cite{rohrbach2015dataset} & \textbf{2015}  &\textbf{94} & \textbf{68,337}  & \textbf{68,375} \\
			\textbf{M-VAD}$^{**}$ \cite{rohrbach2017movie} & \textbf{2015} & \textbf{92} & \textbf{48,986} & \textbf{55,904}  \\
			\textbf{Charades}$^{*}$ \cite{sigurdsson2016hollywood} & \textbf{2016} & \textbf{9,848} & \textbf{9,848} & \textbf{27,847}\\
			\textbf{MSR-VTT} \cite{xu2016msr} & \textbf{2016} & \textbf{7,180} & \textbf{10,000} & \textbf{200,000} \\
			LSMDC $^{**}$\cite{rohrbach2017movie} & 2016 & 128K & 128K & 128K \\
			\textbf{ActivityNet Captions}$^{*}$ \cite{krishna2017dense} & \textbf{2017} & \textbf{20,000} & \textbf{100,000} & \textbf{100,000}\\
			VAS \cite{yu2017supervising}  & 2017 & 144 & 144 & 4032 \\
			YouCook2 $^{*}$\cite{zhou2018towards} & 2018 & 2,000 & 2,000 & 13,829\\
			FSN \cite{yu2018fine} & 2018 & 2,000 & 2,000 & 6,520 \\
			VATEX  $^{*}$\cite{wang2019vatex}  & 2019 & 41,250 & 41,250 & 825,000 \\
			SVCDV \cite{qi2019sports}  & 2019 & 55 & 4,830 & 44,436 \\
			Object-oriented captions \cite{zhu2020understanding} &2020& 75 & 75 & 534 \\
			LIRIS, D2 with descriptions \cite{inacio2021osvidcap}  & 2021 & 167 & 167 &367\\ 
			\hline
		\end{tabular}
	}
\end{table}

\subsubsection{Performance Metrics}
\label{sec:performancemetrics}
As in the image captioning task, the metrics commonly used to evaluate the performance of video captioning are BLEU, METEOR, ROUGE-L, and CIDEr. These metrics are used to evaluate the performance between the system's results against a human interpretation.
The main metrics are described in the following sections.

\textbf{BLEU - Bilingual Evaluation Understudy}.
BLEU~\cite{papineni2002bleu} is one of the most used metrics to evaluate video captioning task. BLEU analyzes co-occurrences of n-grams between the candidate and reference sentences. These matches are positionally independent, and hence the more matches, the better the result. The n-gram is a sequence of $n$ elements; in this case, the elements are the words (or tokens if previous tokenization is performed). Hence, we can calculate BLEU-1 splitting in unigrams, BLEU-2 for bigrams, and so on. 

The clipped n-gram counts for all the candidate sentences and divides by the number of candidate n-grams in the test corpus to calculate the precision score $p_n$.
To calculate several results of several configurations of n-grams, BLEU uses the geometric mean, and to avoid shorter candidates, BLEU applies a penalty factor. The final calculation is performed as:
	\begin{equation}
		\text{BLEU} = BP*exp \left(\sum_{n=1}^{N} w_n \log p_n \right)\,\, ,
	\end{equation}

	\noindent where $p_n$ is the precision score, $BP$ is the brevity penalty, $N$ is the longest sequence of n-grams to consider, and $w_n$ are the weights for each different n-gram length.\footnote{The original paper~\cite{papineni2002bleu} uses $N=4$ and uniform weights, $w_n = 1/N$.} BLEU values range from 0 to 1, being 0 worst case and 1 best possible value.
	For more details about BLEU calculation, please see~\cite{papineni2002bleu} reference.


	\textbf{METEOR - Metric for Evaluation of Translation with Explicit ORdering}. METEOR \cite{banerjee2005meteor} arose to manage the drawbacks in BLEU. The score results of matching words to make a relation translation; this can measure the satisfactorily ordered in the translation.

	The score of METEOR considers an $F_{mean}$ and a $Penalty$. $F_{mean}$ is based on precision ($P$) and recall ($R$) for unigram matches, and it is considered a harmonic mean. The Equation \ref{eq_fmean} shows this mathematic relation.

	\begin{equation}\label{eq_fmean}
		F_{mean}=\frac{10PR}{R+9P}
	\end{equation}

	Specifically, $P$ is the ratio of the unigrams of reference and the unigrams obtained for the model, and $R$ is the ratio of the mapped unigrams of the reference to the total.

	The final consideration is the $Penalty$; it consists in mapping the fewest feasible quantity of chunks in the sentence to unigrams; if the chunks increase, the penalty does the same. The equation \ref{eq:penallty} expresses this relation.

	\begin{equation}\label{eq:penallty}
		Penalty=0.5*\left ( \frac{no.\, chunks}{no.\, unigrams\_matched} \right )^{3}
	\end{equation}

	Finally, the METEOR score is computed as the equation \ref{eq:meteor} shows.

	\begin{equation}\label{eq:meteor}
		\text{METEOR} = F_{mean}*(1-Penalty)
	\end{equation}

	\textbf{ROUGE - Recall-Oriented Understudy for Gisting Evaluation}. ROUGE \cite{lin2004rouge} is a metric evaluation that measures the number of overlapping in n-gram, term sequences, and word pairs for the captions generated. This metric has four versions:  ROUGE-N, ROUGE-L, ROUGE-W, and ROUGE-S. ROUGE-L is used in the video captioning task, and it estimates the longest common subsequence (LCS). The first step is to obtain the Recall ($R_{lcs}$) and Precision ($P_{lcs}$) associated with LCS.

	$R_{lcs}$ is the ratio of the maximum length of reference ($s_r$) and predicted ($s_p$) sentence $LCS(s_r,s_p)$, and the total words of the reference sentence (see equation \ref{eq:rlcs}). $P_{lcs}$ is very similar to the recall, but the length consideration is related to the predicted sentence (see equation \ref{eq:plcs}).

	\begin{equation}\label{eq:rlcs}
		R_{lcs}=\frac{LCS(s{_{r}},s{_{p}})}{len(s{_{r}})}\,\, ,
	\end{equation}

	\begin{equation}\label{eq:plcs}
		P_{lcs}=\frac{LCS(s{_{r}},s{_{p}})}{len(s{_{p}})}\,\, .
	\end{equation}

	Finally, the ROUGE-L score is calculated by the equation \ref{eq:rougel}:

	\begin{equation}\label{eq:rougel}
		\text{ROUGE-L}=\frac{(1+\beta^{2})R_{lcs}P_{lcs}}{R_{lcs}+\beta^{2}P_{lcs}}\,\, ,
	\end{equation}
	where $\beta$ is a constant number to consider the precision.

	\textbf{CIDEr - Consensus-based Image Description Evaluation}. CIDEr measures the similarity of the generated captions versus a set of ground truth sentences created by humans~\cite{vedantam2015cider}. CIDER arose as an answer to the necessity for a metric that could better correlate with human judgment, employing a {\it consensus} between the several proposed captions, or references, in a dataset. In this case, one can have two generated sentences called B and C, and a reference sentence called A. CIDEr tries to solve the question of which of the two sentences, B or C, is more similar to A?.

	To evaluate the generated caption (or candidate sentence $c_i$) for an image $I_i$ versus a set of image descriptions or references $S_i ={s_{il},...,s_{im}}$, with CIDEr, firstly, all words are converted into its root form. Each sentence is represented using the n-grams approach, originally in~\cite{vedantam2015cider}\footnote{Only values of 1 to 4 for $n$ are considered.}. The n-gram $w_k$ is a set of one or more sequential words. The TF-IDF approach is used to generate CIDEr metric. See~\cite{robertson2004understanding} for TF-IDF details. 



	The CIDEr$_n$ for n-grams of size $n$ is calculated with the average cosine similarity between the candidate caption and the references as follows:
	\begin{equation}
		\text{CIDEr}_n(c_i, S_i) = \frac{1}{m} \sum_{j}{}\frac{g^n(c_i) \cdot g^n(s_{ij})}{ ||  g^n(c_i) || \, ||  g^n(s_{ij}) ||}\,\, ,
	\end{equation}
	where $g^n(c_i)$ is a vector formed by $g_k(c_i)$ corresponding to all the n-grams of size n, and $||  g^n(c_i) ||$ is the magnitude of the vector $g^n(c_i)$, is the same for $g^n(s_{ij})$.
	Finally, CIDER combines all the results from $n$ values for n-grams as follows:
	\begin{equation}
		\text{CIDEr}\,(c_i, S_i) = \sum_{n=1}^{N} w_n\, \text{CIDEr}_n(c_i, S_i) \,\, ,
	\end{equation}

	\subsubsection{Performance Metrics Examples}

	To exemplify, Table \ref{tab:metric_example} shows a list of the different values that can be calculated for a single predicted caption, given five different human-annotated captions references. This sample sentence was extracted from the MS-COCO dataset \cite{lin2014microsoft} for image annotation, but the same procedure can be extrapolated to video captioning.

	\begin{table}[h]
		\caption{Different values of metrics for a single proposed caption considering five references.}
		\label{tab:metric_example}
		\centering
		\resizebox{9cm}{!}
		{
			\begin{tabular}{l|l}
				\textbf{Predicted caption} & \textbf{References}\\
				& Four donuts in a box with a variety of frostings.\\
				A box of different types of & A close up of many different kinds of doughnuts.\\
				doughnuts on a table. & A group of donuts sitting in a box.\\
				& An image of a box of donuts in a box.\\
				& A group of different types of doughnuts in a box\\
				\vspace{0.5cm}
			\end{tabular}
		}
		\footnotesize{
			\begin{tabular}{|c|c|}
				\hline
				Metric  & Value    \\ \hline
				BLEU-1~\cite{papineni2002bleu} & 0.800000\\
				BLEU-2~\cite{papineni2002bleu} & 0.730297\\
				BLEU-3~\cite{papineni2002bleu} & 0.643660 \\
				BLEU-4~\cite{papineni2002bleu} & 0.525382 \\
				METEOR \cite{banerjee2005meteor} & 0.405500 \\
				ROUGE-L \cite{lin2004rouge} & 0.600000\\
				CIDEr~\cite{vedantam2015cider} & 2.203552\\
				\hline
		\end{tabular}}

	\end{table}

	As can be seen, the same predicted caption could have very different values regarding the metric employed.


	\section{Results and discussion}
	\label{sec:resultsandiscussion}

	The presented results are based on the review and analysis of 105 published papers in journals and conferences related to video captioning; additional reviews papers were also consulted. The range of years of publication of the reviewed papers is from 2016 to 2021.
	Our review was done by selecting specific points from all papers, these points are title, year, method' description, datasets used, results reached, and where the paper was published. We did the analysis detailed in the following sections from all this information.

	From the 105 papers, we counted 20 different datasets reported; some datasets were subsets from others. Please see section \ref{sec:datasets} for more details about the descriptions of each dataset reviewed.

\begin{figure*}[!h]
		\centering
		{\includegraphics[width=0.75\textwidth]{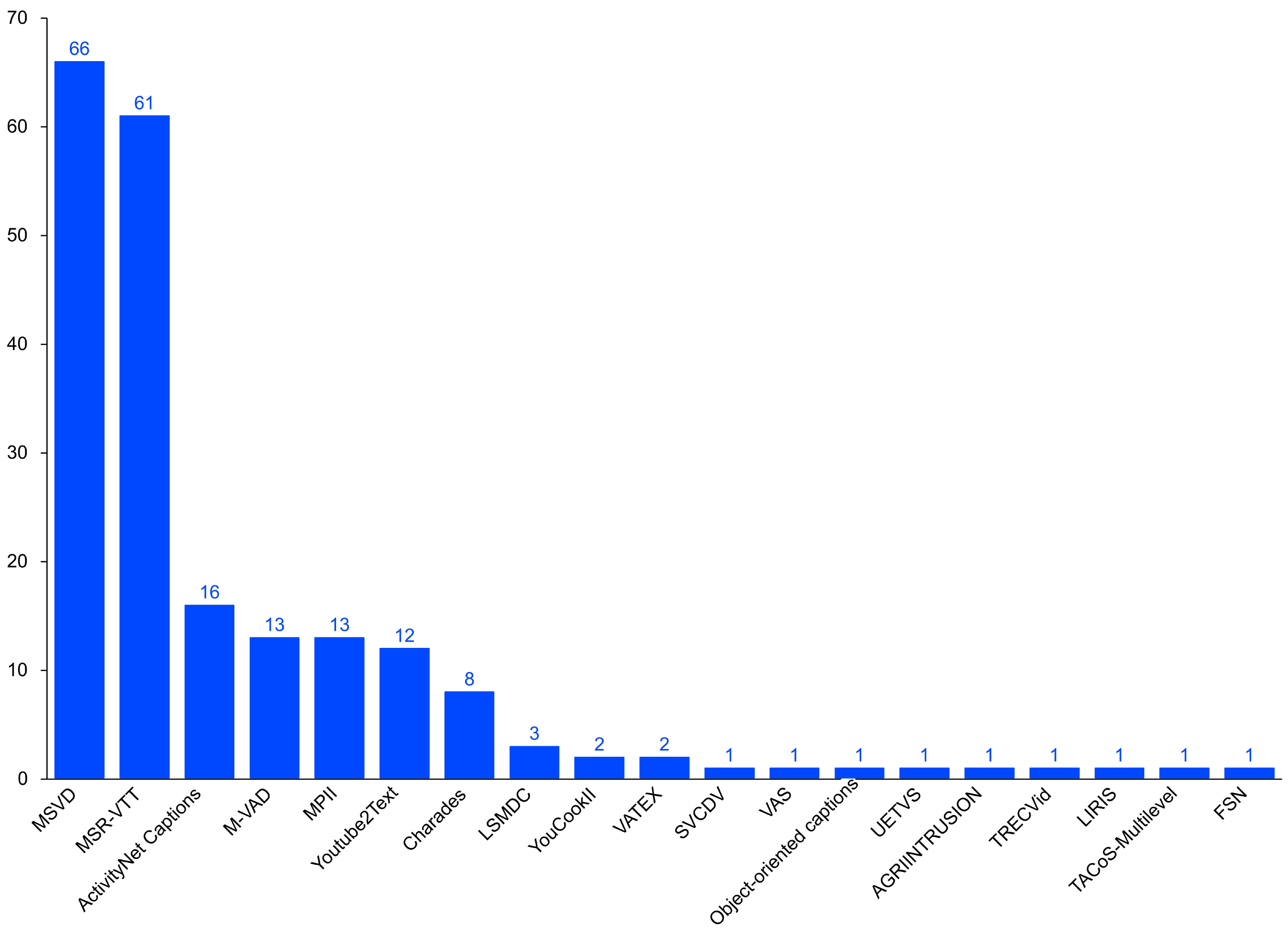}}
		\vspace{-0.2cm}
		\centering \caption{Number of citations for each dataset}
		\label{fig:chart-BD-all}
	\end{figure*}

	Figure~\ref{fig:chart-BD-all} shows a histogram of all the datasets found in our literature review.
	Here, it can be seen that the most cited dataset is MSVD, with 66 times used in different papers. The second is MSR-VTT, used in 61 different works. It can be observed that these two datasets are, by far, the most utilized. Nevertheless, there is a block of five datasets with several citations higher than or equal to 8, and there are also many datasets with few cites; exactly 13 datasets were referenced in less than four works.

\begin{figure}[h]
		\centering
		{\includegraphics[width=0.5\textwidth]{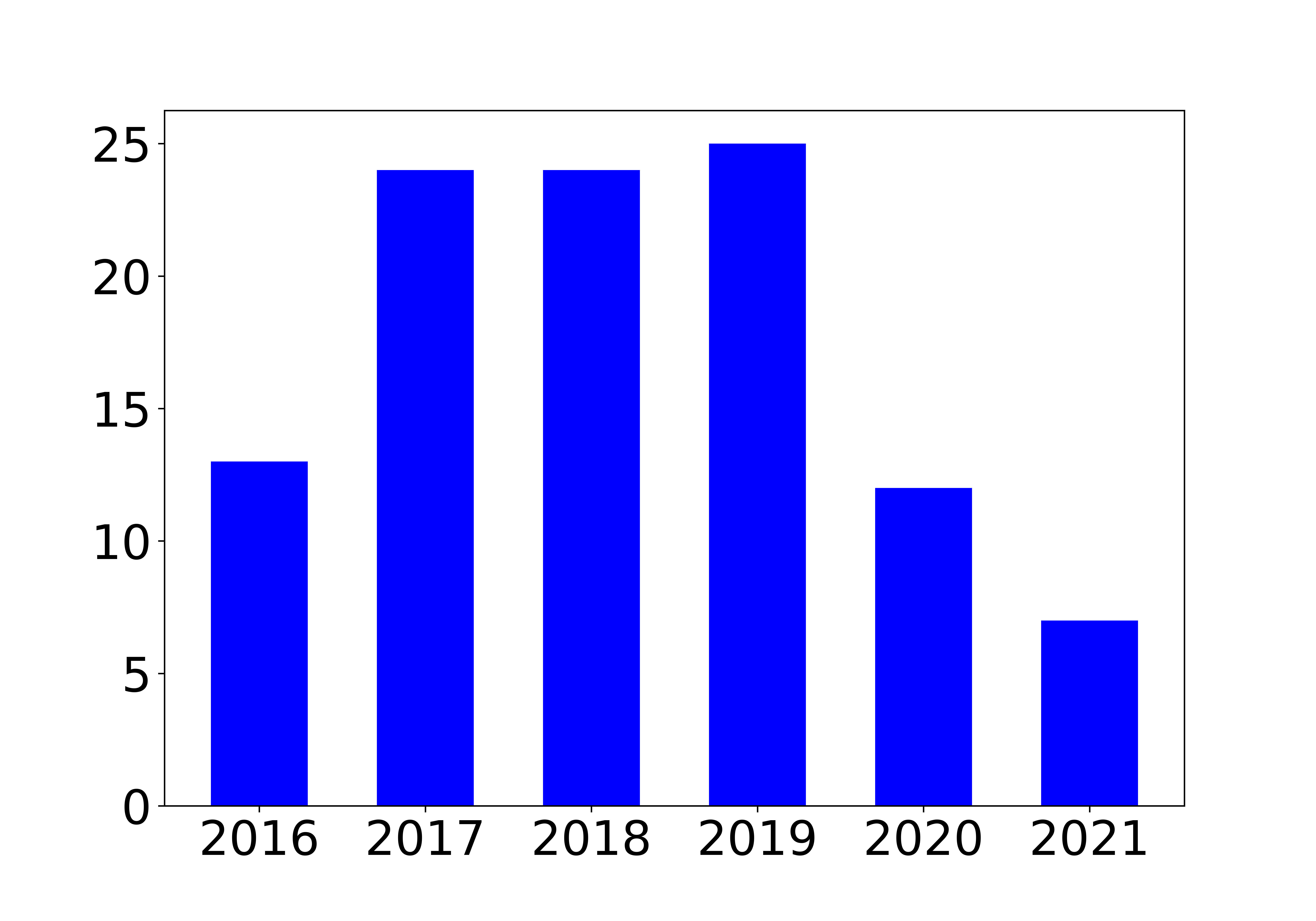}}
		\caption{Number of papers related to video captioning task published by year, since 2016 to 2021}
		\label{fig:allpaperstime}
	\end{figure}

	As mentioned, the analysis was done from 2016 to 2021, and to observe how many manuscripts were published over the years,  Figure~\ref{fig:allpaperstime} shows the number of publications per year being 2019, the year with more papers published related to the video captioning task. In 2020 and 2021, only 12 and 7 manuscripts, were published, respectively.
	This could be due to the COVID-19 pandemic since more than 23 papers were published the previous year.

\begin{figure}
		\centering
		{\includegraphics[width=.35\textwidth]{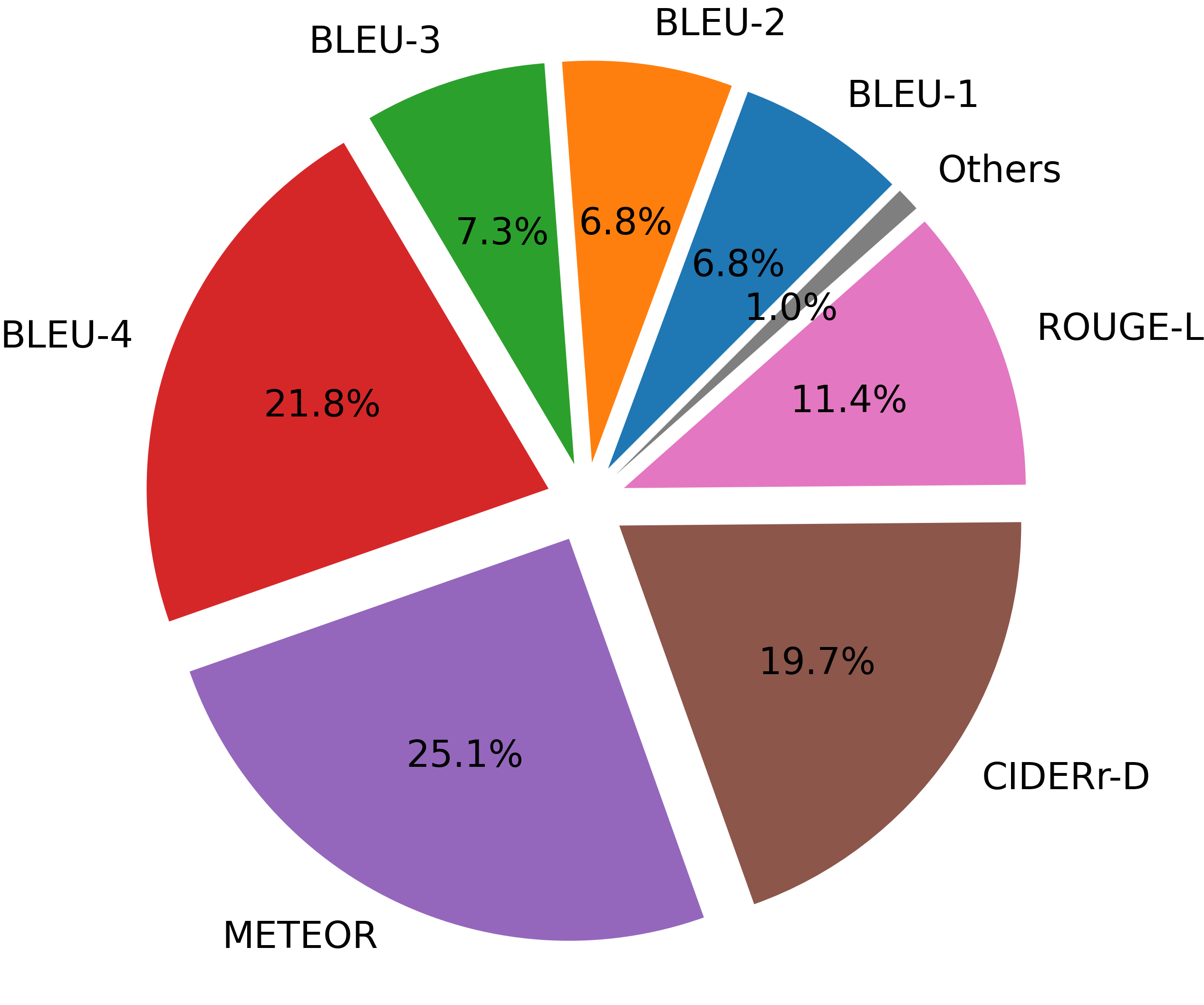}}
		\caption{Metrics usage over 105 papers revised}
		\label{fig:metricas}
	\end{figure}

	To follow a simple analysis scheme, first, the frequency of usage of all the performance metrics was calculated. As a result, Figure~\ref{fig:metricas} displays the percentage of usage of each reported metric. Here, it can be seen that the most used are METEOR, followed by BLEU-4 and CIDERr-D, with second and third place, respectively.
	BLEU-1, BLEU-2, and BLEU-3 have a similar frequency, and in the label ``Other'' we joined those metrics with less than two reported results; these are the Average-Recall (AR), SPICE, FCE,  RE,  and Self-BLEU.

	To analyze in an organized way, we sorted the results in each dataset based on the METEOR metric, which is the most used (see Figure~\ref{fig:metricas}), at least in the reviewed literature.
	After the results were sorted, we took the top five in each dataset. Table~\ref{tab:resultadostop5} shows the best five results reported in the seven most-used datasets. Here, is observed that the best ranked methods are not necessarily the most recent ones. The best result on the MSVD dataset was reached by~\cite{perez2021improving} on the three compared metrics. The same for the MSR-VTT dataset, where Gao et al.~\cite{gao2019hierarchical} achieved the best result in all metrics. Nevertheless, this is not the case for the ActivityNet Captions dataset, where~\cite{xiong2018move} obtained the best result in the METEOR and BLEU-4 metrics and the worst result on CIDERr-D.
	Only the METOR metric was used to report results for the M-VAD dataset because there were missing values in both BLEU-4 and CIDERr-D. The best result on the MPII dataset was achieved by~\cite{pan2017video}, for the Youtube2Text dataset the best result on the METEOR and CIDERr-D was reached by~\cite{chen2018tvt}, but this was not the case for BLEU-4.
	Finally, on the Charades dataset, \cite{zhao2019cam} obtained the best score if we only consider the METEOR metric, but for BLEU-4 and CIDERr-D, that is not the case.


	From these reported results, we can observe that there is no simple way to choose the best method because we have multiple metrics. Another argument for this, is that most of the reviewed manuscripts usually employ just one or two datasets in their evaluation, making it difficult to declare an overall winner.

	\begin{table}[!h]

		\caption{{The best five results according to METEOR on the seven most used datasets}}
		\label{tab:resultadostop5}
		\resizebox{9cm}{!}
		{
			\begin{tabular} { cccccc }
				\toprule
				\textbf{MSVD} &   Year	 & Work	&   METEOR &    BLEU-4 &	 	CIDERr-D\\
				\toprule
				&   \bf{2021} &	\cite{perez2021improving} & 	\bf{41.90}   &	64.40   &	111.50 \\
				&	2018  &	\cite{pu2018adaptive}     &	    38.03   &	54.27   &	78.31 \\
				&	2020    &	\cite{pan2020spatio}	  & 	36.90	&   52.20   &	93.00 \\
				&	2021	&\cite{chen2021motion}     &	    36.90   &	55.80   &	74.50	\\
				& 2020 &	\cite{zhang2020object}	      &	    36.40   &	54.30   &	95.20 \\
				\toprule
				\\
				\textbf{MSR-VTT} &   Year	 & Work	&   METEOR &    BLEU-4 &	 	CIDERr-D\\
				\toprule
				&  \bf{2019} &	\cite{gao2019hierarchical}      &  \bf{33.50}   &	54.30   &  72.80 \\
				&   2021 &	\cite{perez2021improving}       &	30.40   &	46.40   &	51.90   \\
				&	2018 &	\cite{pu2018adaptive}	        &	29.98   &	45.01   &	51.41	\\
				&	2019 &	\cite{hou2019joint}             &	29.70   &	42.30   &	49.10	\\
				&	2019 &	\cite{chen2019generating}       &	29.61   &	44.91   &	51.80	\\

				\toprule

				\\
				\textbf{ActivityNet Captions} &   Year	 & Work	&   METEOR &    BLEU-4 &	 	CIDERr-D\\
				\toprule

				&   \bf{2018}     &\cite{xiong2018move}	    &	\bf{14.75}   &	8.45    &	14.15	\\
				&	2019	&\cite{mun2019streamlined}	&	13.07   &	1.28    &	43.48	\\
				&   2020	&\cite{iashin2020multi}	    &	11.72   &	2.86    &	NA	\\
				&	2019	&\cite{hou2019joint}        &   11.30   &	1.90    &	44.20 \\
				&	2019	&\cite{zhang2019show}       &	10.71	&   1.64    &	31.41	\\

				\toprule

				\\
				\textbf{M-VAD} &   Year	 & Work	&   METEOR &    BLEU-4 &	 	CIDERr-D\\
				\toprule

				&   \bf{2017}       &	\cite{pasunuru2017multi} &	\bf{7.40}    &	NA    &	NA \\
				&	2017    &	\cite{pan2017video}	    &	7.40	&   NA     &	NA	 \\
				&	2017    &	\cite{baraldi2017hierarchical}	&	7.30        &	NA     &	NA	\\
				&	2018    &	\cite{xu2018sequential}	&   7.20    &	NA     &   	NA	 \\
				&	2018    &	\cite{pu2018adaptive}	&	7.12    &	2.08    &	9.14	\\
				\toprule

				\textbf{MPII} &   Year	 & Work	&   METEOR &    BLEU-4 &	 	CIDERr-D\\
				\toprule
				&   \bf{2017}    &	\cite{pan2017video}	    &	\bf{28.8}    &	40.8    &	47.1 \\
				&	2018    &	\cite{xu2018dual}       &	7.9     &	1.9     &	NA	\\
				&	2019    &	\cite{zhao2019cam}	    &	7.8	    &   NA     &	NA	\\
				&	2019    &	\cite{xu2019multi}	    &   7.7     &	0.8     &	NA	\\
				&	2016    &	\cite{pan2016jointly}	&	7.3     &	NA     &	NA \\
				\toprule

				\textbf{Y2T} &   Year	 & Work	&   METEOR &    BLEU-4 &	 	CIDERr-D\\
				\toprule
				&   \bf{2018}    &	\cite{chen2018tvt}	        &  \bf{35.23}   &	53.21  & 86.76	\\
				&	2018	&   \cite{zolfaghari2018eco}	&	35.00   &	53.50	&    85.80	\\
				&	2020	&   \cite{wei2020exploiting}    &	34.40   &	46.80   &	85.70	\\
				&	2017	&   \cite{hori2017attention}	&	34.30   &	56.80   &	72.40	\\
				&	2017	&   \cite{chen2017video}        &	34.21   &	47.56   &	79.57 \\
				\toprule

				\textbf{Charades} &   Year	 & Work	&   METEOR &    BLEU-4 &	 	CIDERr-D\\
				\toprule
				&   \bf{2019}    &	\cite{zhao2019cam}      &	\bf{19.7}    &	12.9	&   18.8 \\
				&	2017    &	\cite{li2017mam}	    &	19.1	&   12.7	&   18.3	\\
				&	2018    &	\cite{zhao2018video}	&	19.0	&   13.3	&   18.0	\\
				&	2018    &	\cite{wang2018video}    &	18.7	&   18.8	&   23.6	\\
				&	2019    &	\cite{hu2019hierarchical}&	18.4	&   14.5	&   23.7	\\
				\toprule

				\bottomrule
			\end{tabular}
		}
	\end{table}

\begin{figure*}[!h]
		\begin{tabular}{ccc}
			\includegraphics[width=55mm]{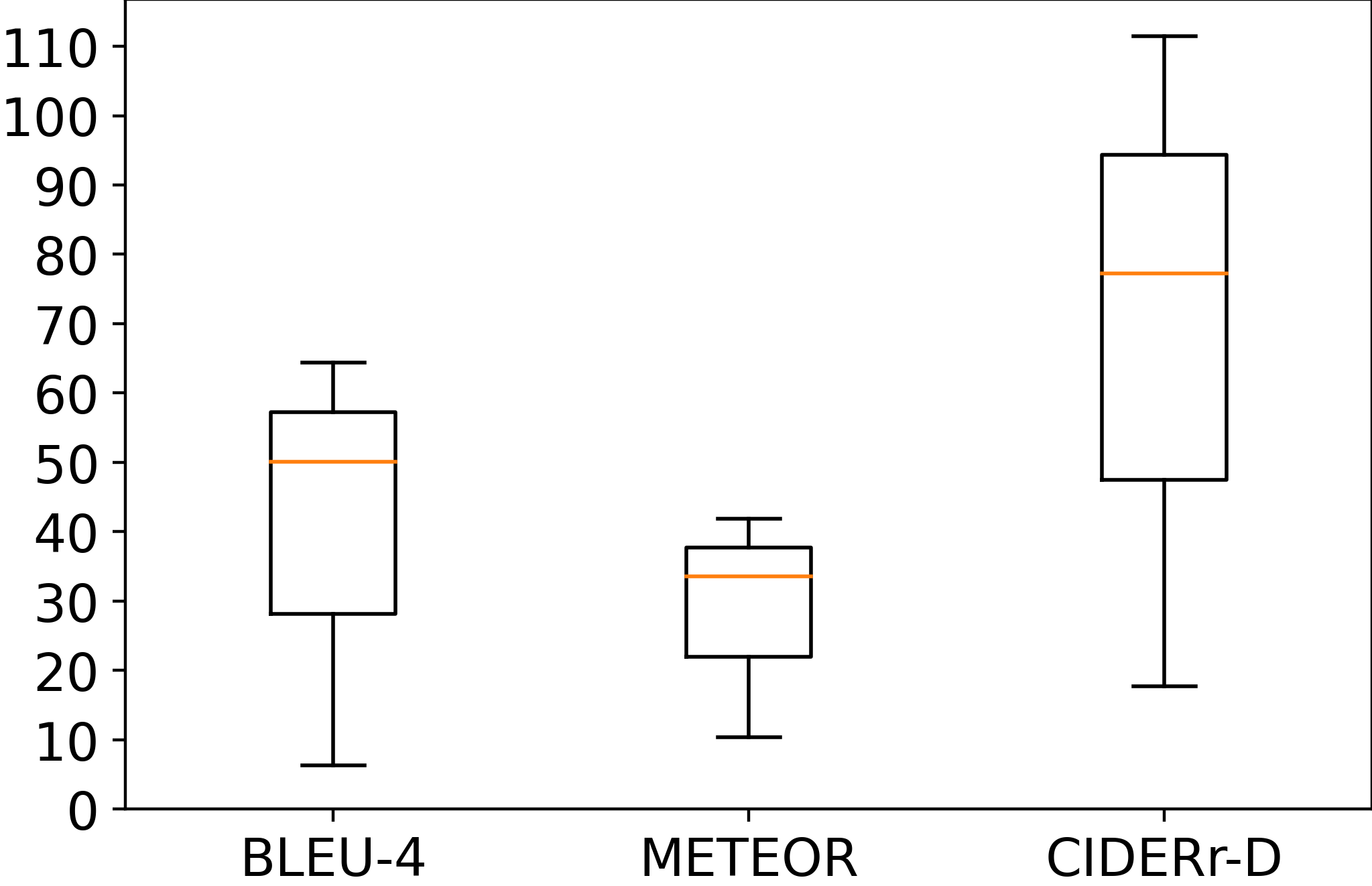} &   \includegraphics[width=55mm]{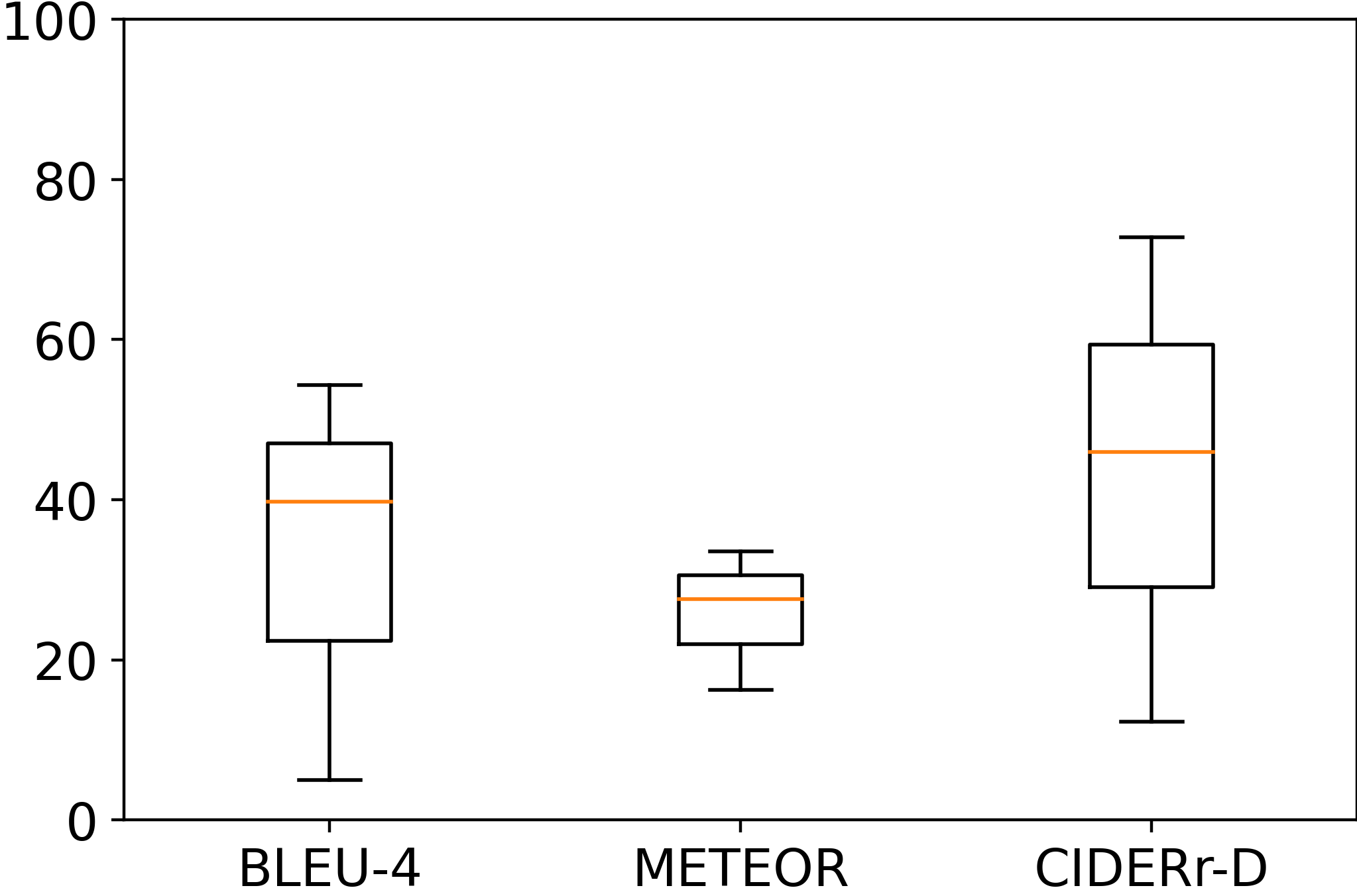}&   \includegraphics[width=55mm]{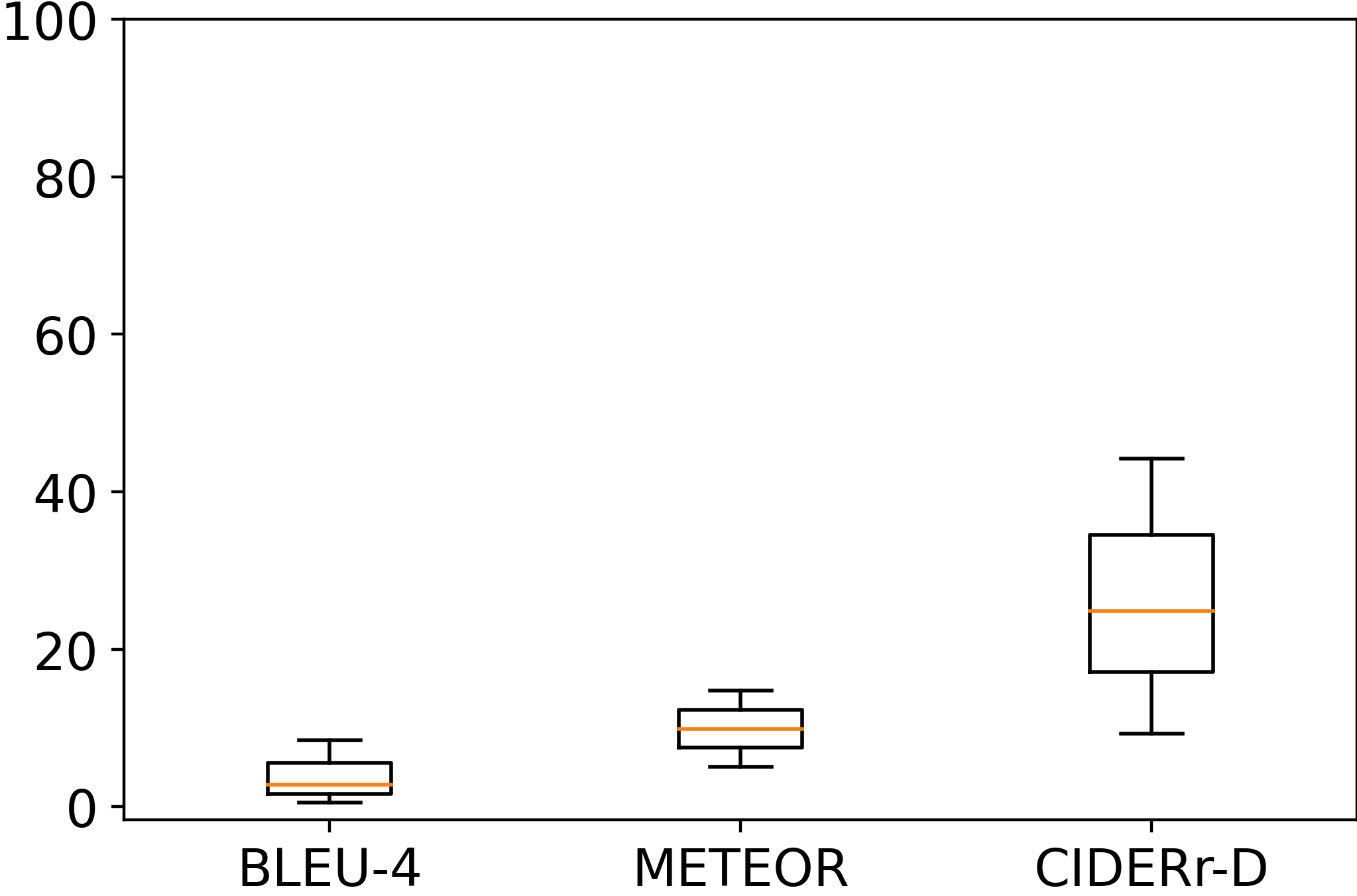} \\
			(a) MSVD  & (b) MSR-VTT & (c) ActivityNet Captions \\[6pt]
			\includegraphics[width=55mm]{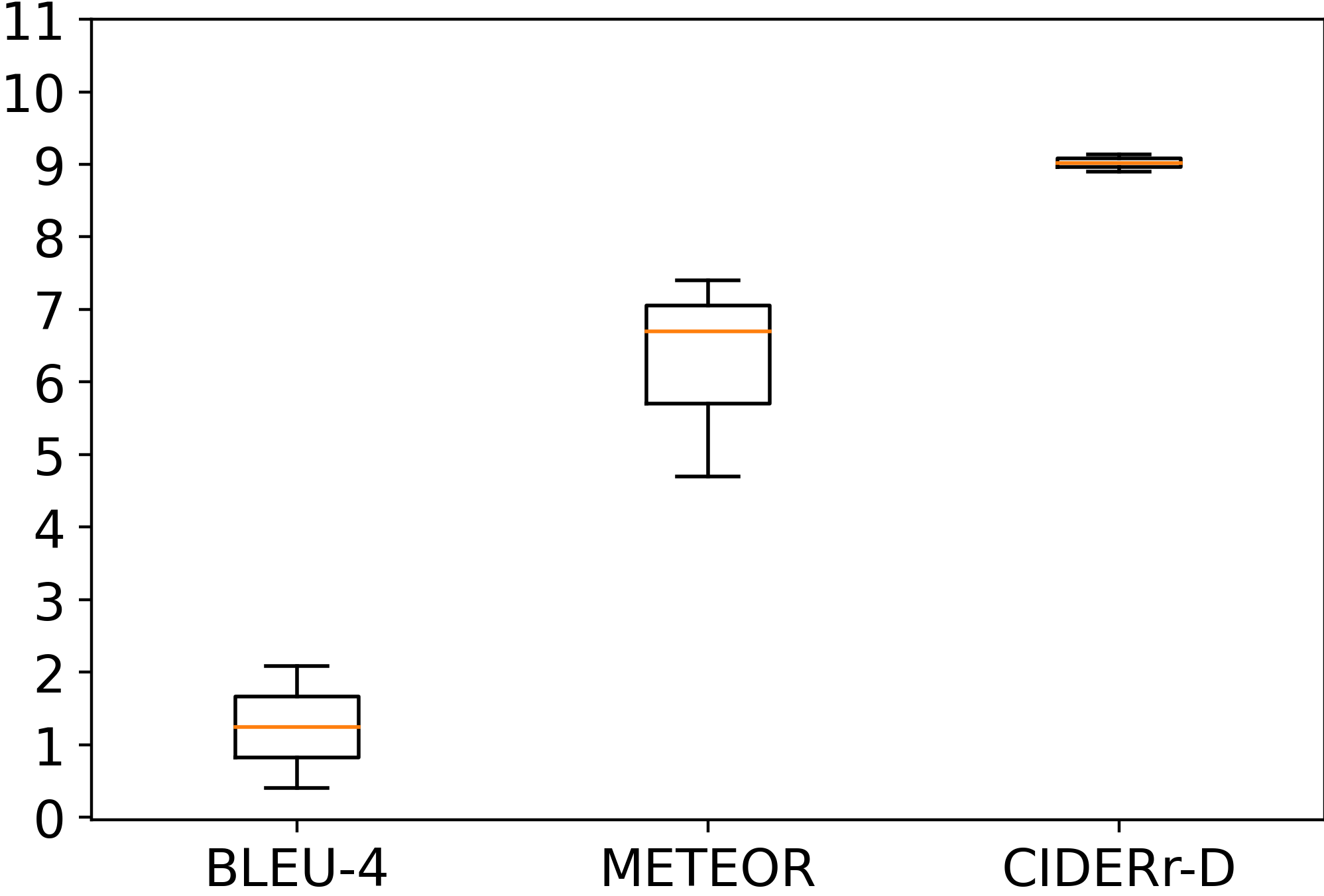} &   \includegraphics[width=55mm]{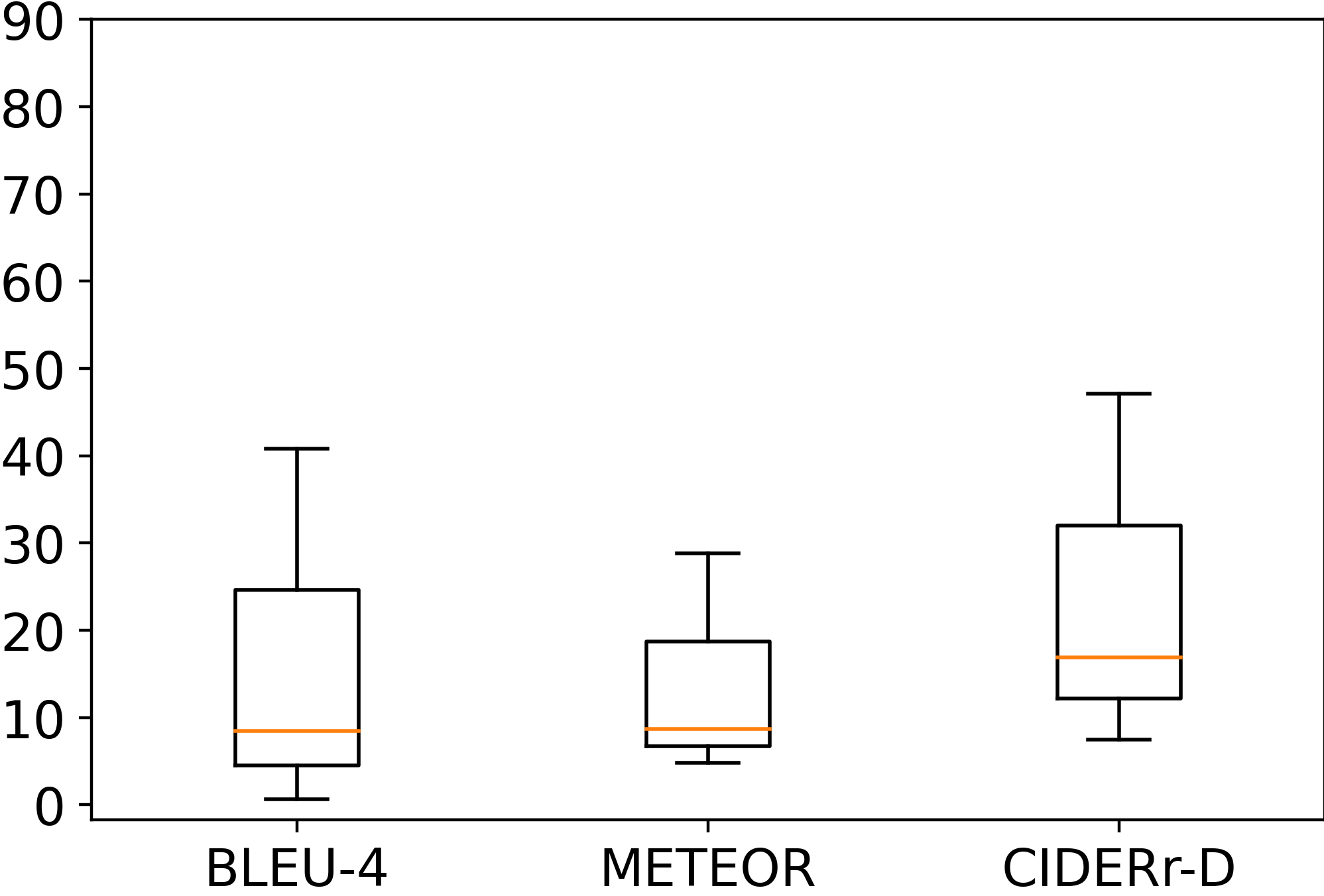}&   \includegraphics[width=55mm]{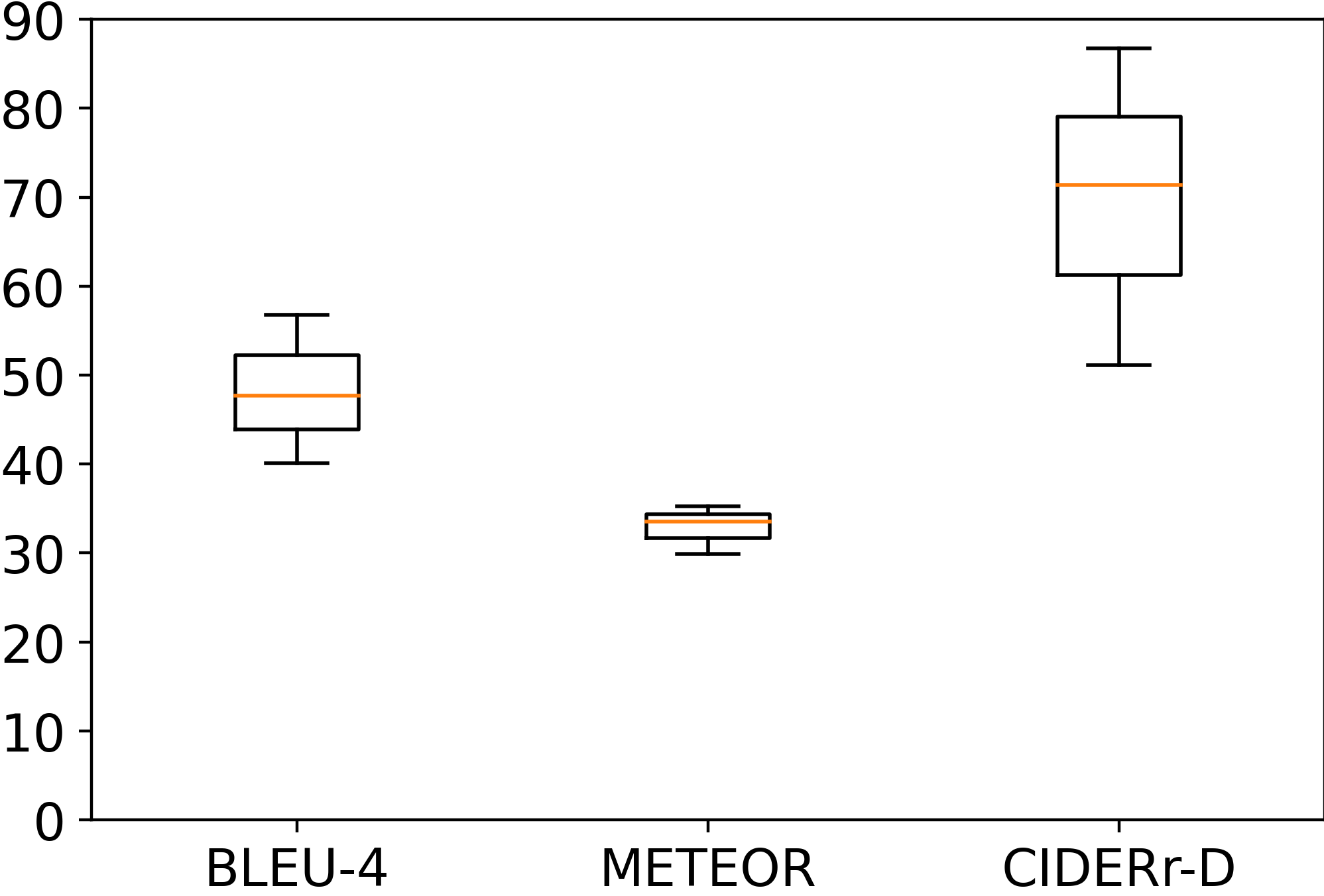} \\
			(c) M-VAD & (d) MPII & (c) Youtube2Text\\[6pt]
			\multicolumn{3}{c}{\includegraphics[width=55mm]{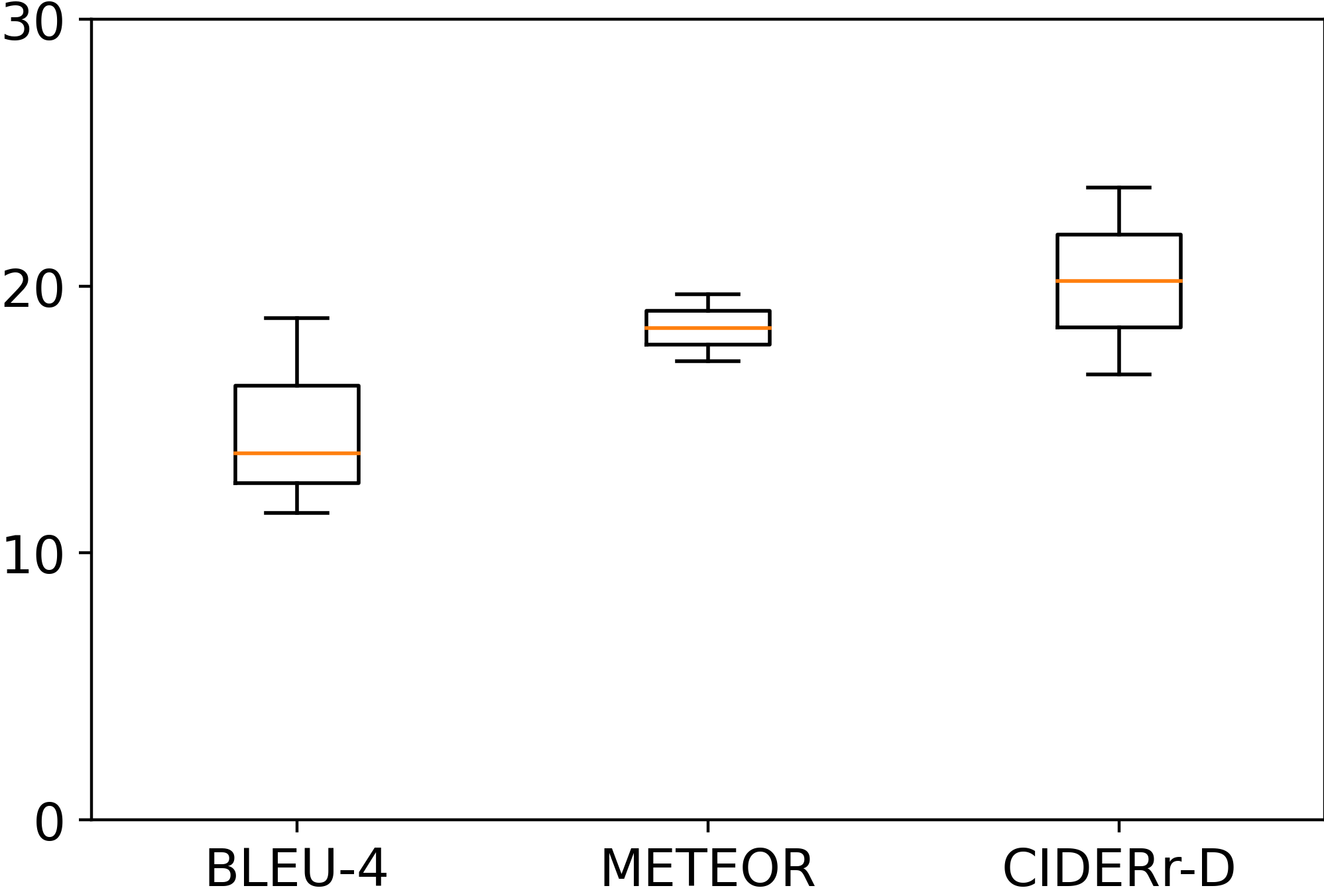} }\\
			\multicolumn{3}{c}{(e) Charades}
		\end{tabular}
		\caption{Results in most used datasets with three most used metrics}
		\label{fig:allresults}
	\end{figure*}

	For these reasons, we present a set of box plots to observe the behavior of each of these three most used metrics over all the datasets. Figure~\ref{fig:allresults} shows the behavior of BLEU-4, METEOR, and CIDERr-D per dataset, over MSVD, MSR-VTT, ActivityNet Captions, M-VAD, MPII, Youtube2Text, and Charades datasets, respectively. All the reported results over this revision of 105 works are grouped by dataset and metric. For instance, on the MSVD dataset, we can observe the median, min, and max values on each metric. Using the BLEU-4 metric, the median value is around 50, but most of the works that reported results on BLEU-4 are below this median; the same happens for METEOR, and the CIDERr-D metric. These aspects could be observed in the rest of the box plots from each dataset, and several insights could be noted. For instance, the METEOR metric has less variation on Youtube2Text despite all three metrics having the same amount of reported results. The same happens with the Charades, and ActivityNet Captions datasets. Another outlier behavior is from the CIDERr-D metric on the M-VAD dataset, and this is because (as we can see in Table~\ref{tab:resultadostop5}) only two results were reported using the CIDERr-D metric.

\begin{figure}[h]
		\centering
		\includegraphics[width=85mm]{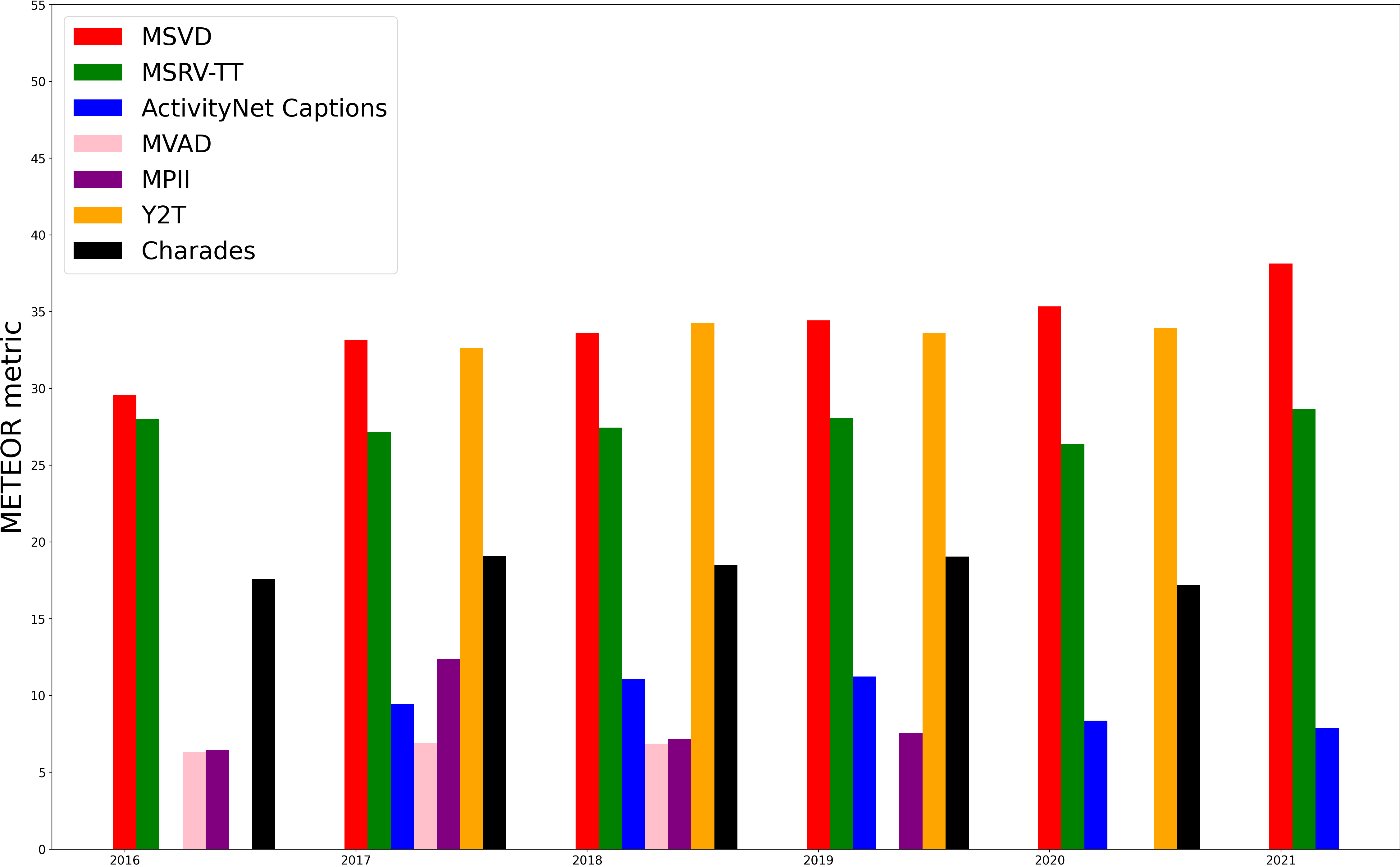}
		\caption{All datasets through years 2016-2021 using METEOR metric}
		\label{fig:allDB-meteor}
	\end{figure}

\begin{figure}[h]
		\centering
		\includegraphics[width=85mm]{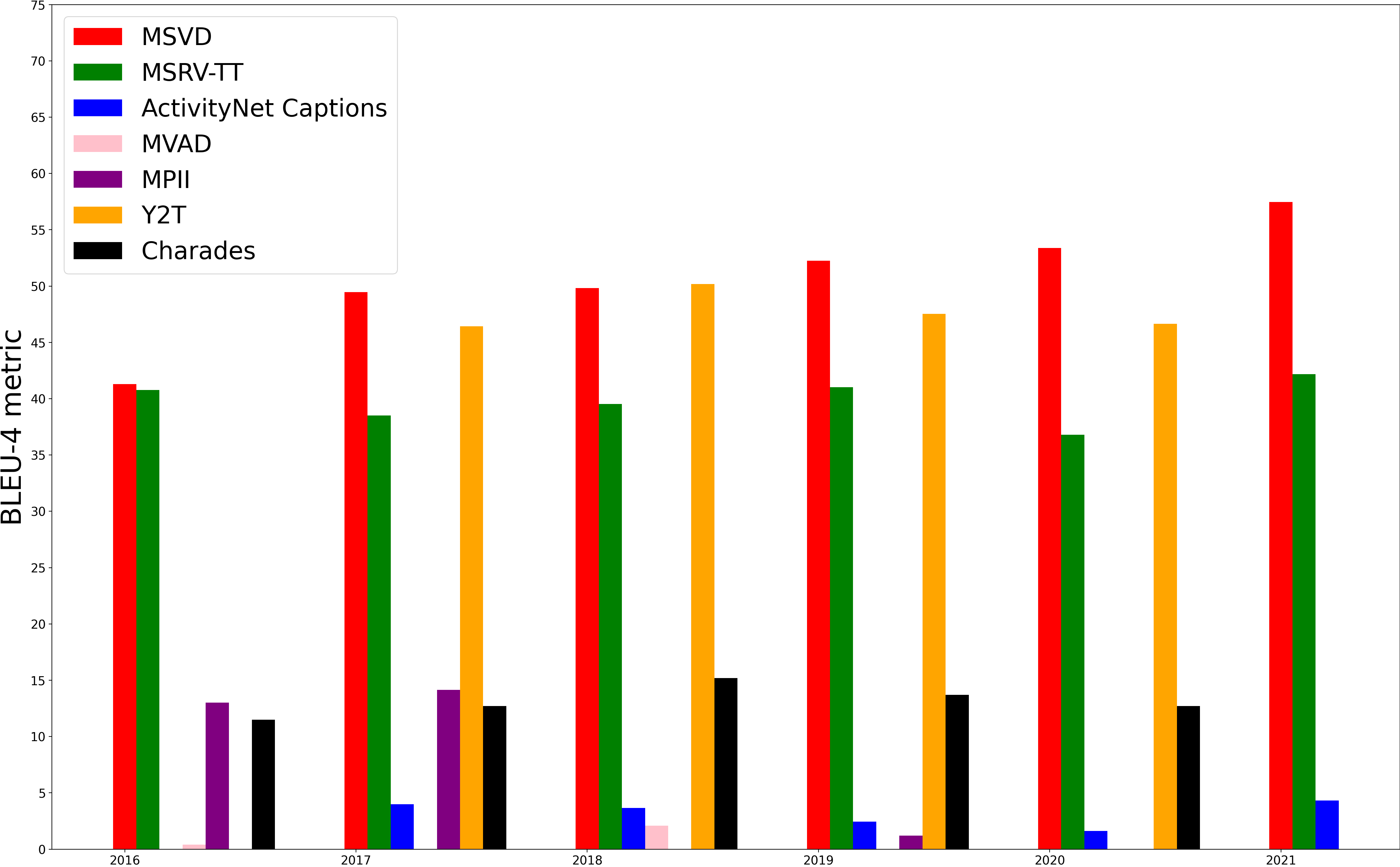}
		\caption{All datasets through years 2016-2021 using BLEU-4 metric}
		\label{fig:allDB-bleu4}
	\end{figure}

\begin{figure}[h]
		\centering
		\includegraphics[width=85mm]{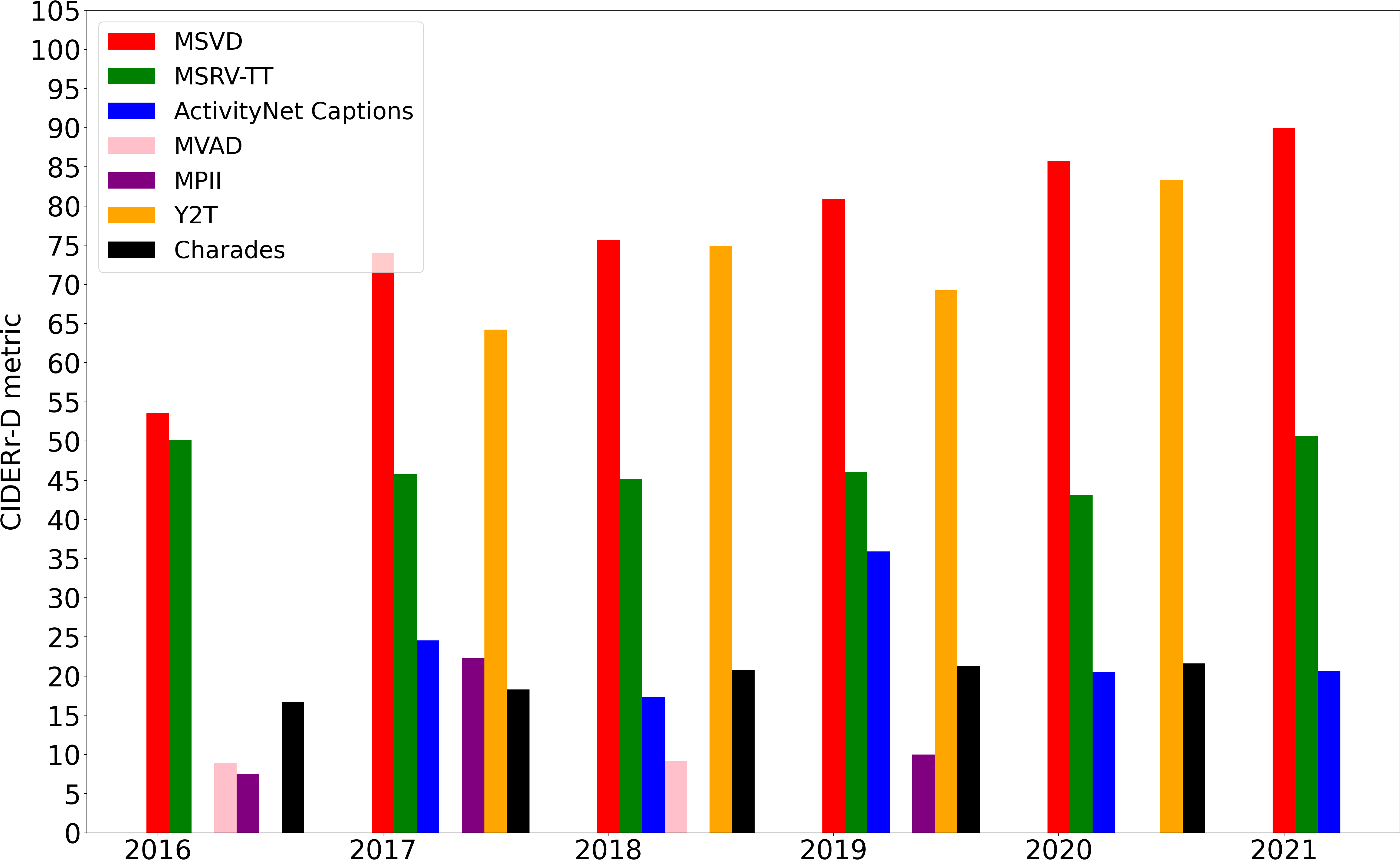}
		\caption{All datasets through years 2016-2021 using CIDERr-D metric}
		\label{fig:allDB-cider}
	\end{figure}

	Trying to present the evolution of all the reported results over the years, in Figures~\ref{fig:allDB-meteor}, \ref{fig:allDB-bleu4}, and \ref{fig:allDB-cider}, we show the best result over the years from each dataset using METEOR, BLEU-4, and CIDERr-D metrics. In Figure~\ref{fig:allDB-meteor}, it can be seen that the highest results always were achieved on the MSVD dataset (red bar in the figure). It is also one of the datasets used in all the studied years, along with MSR-VTT.
	The lowest results on the METEOR metric, were achieved by MVAD and MPII datasets related papers.
	For the BLEU-4 metric, the MSVD dataset got the highest results again over the period studied. Contrary to METEOR, the lowest scores were obtained with ActivityNet Captions and M-VAD datasets related works.
	With the CIDERr-D metric, once again, the MSVD dataset achieved the highest reported results and the lowest on MPII and MVAD datasets.
	As noted, models trained using MSVD reached the best result in 2021, but this is not the case with the other datasets that achieved the best reported results in different years, from 2017, 2018, and 2019.

	In order to obtain a global ranking, we computed individual rankings on the three most used metrics. We gathered the results on METEOR, BLEU-4, and CIDERr-D from all the reviewed works in the seven most-used datasets.
	This ranking procedure considers the top 5 results according to the metric. Analyzing the set of papers, we can compute the result of each of them in the seven most-used datasets. That means, if a paper achieves the first position, 5 points will be assigned to it; the second position grants 4 points, and go on until the fifth position, with only one point; in this way, we get the paper with the highest number of points. Finally, the sum of the points is divided by 7 (total number of datasets used) to normalize them.

\begin{figure*}[!h]
		\begin{tabular}{cc}
			\centering
			\includegraphics[scale=0.13]{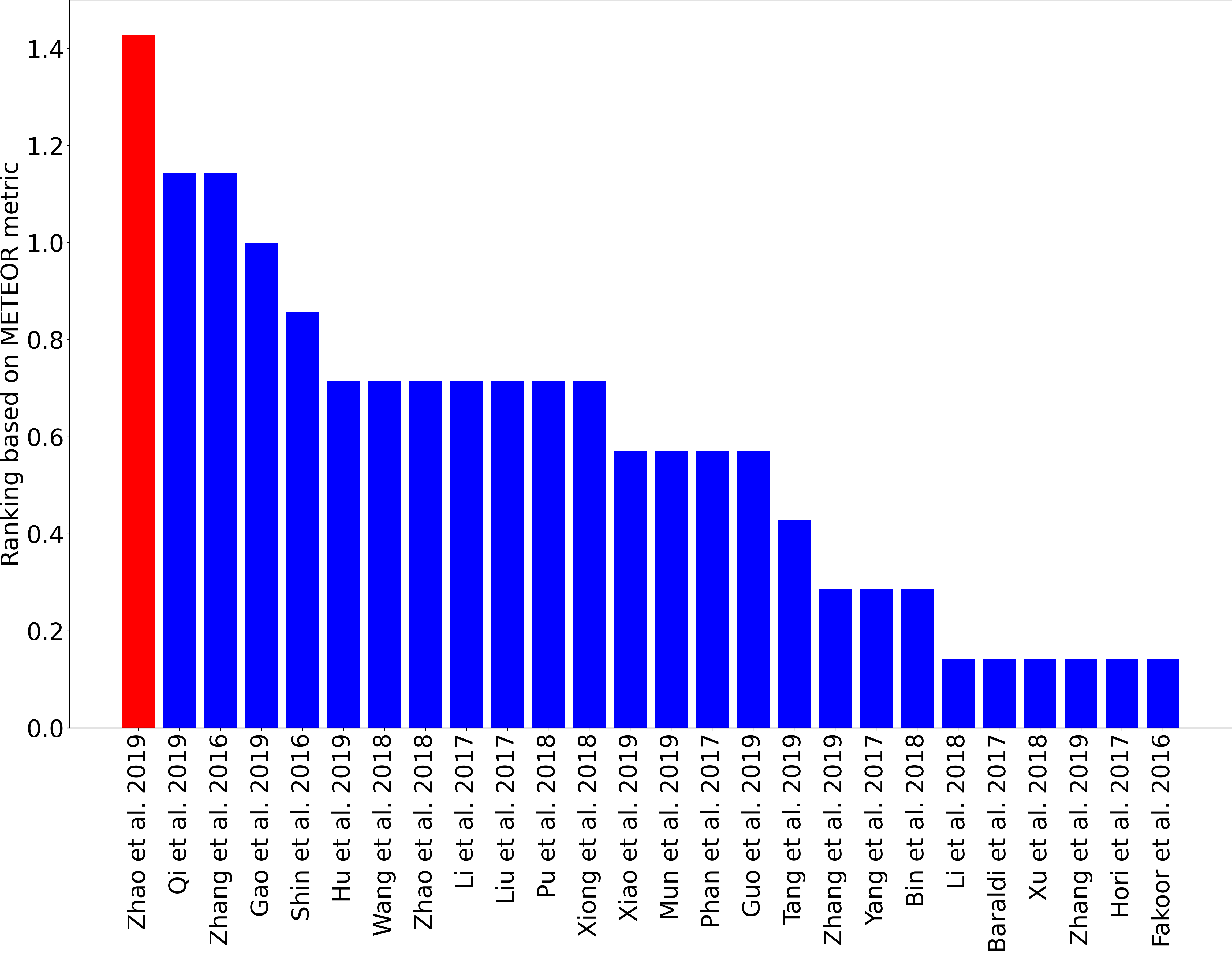} &    \includegraphics[scale=0.13]{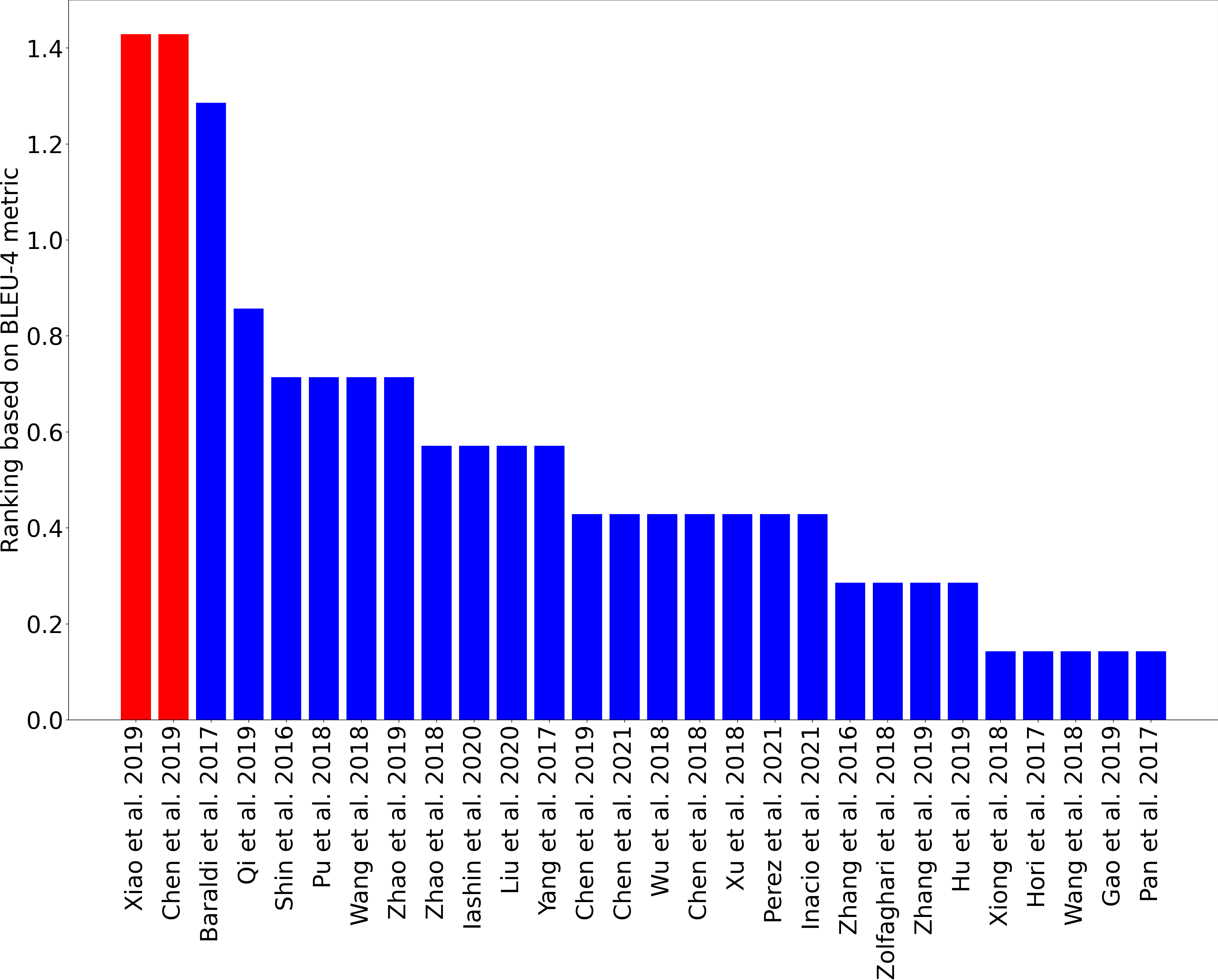} \\
			(a) Ranking based on METEOR  & (b)  Ranking based on BLEU-4  \\[6pt]
			\includegraphics[scale=0.13]{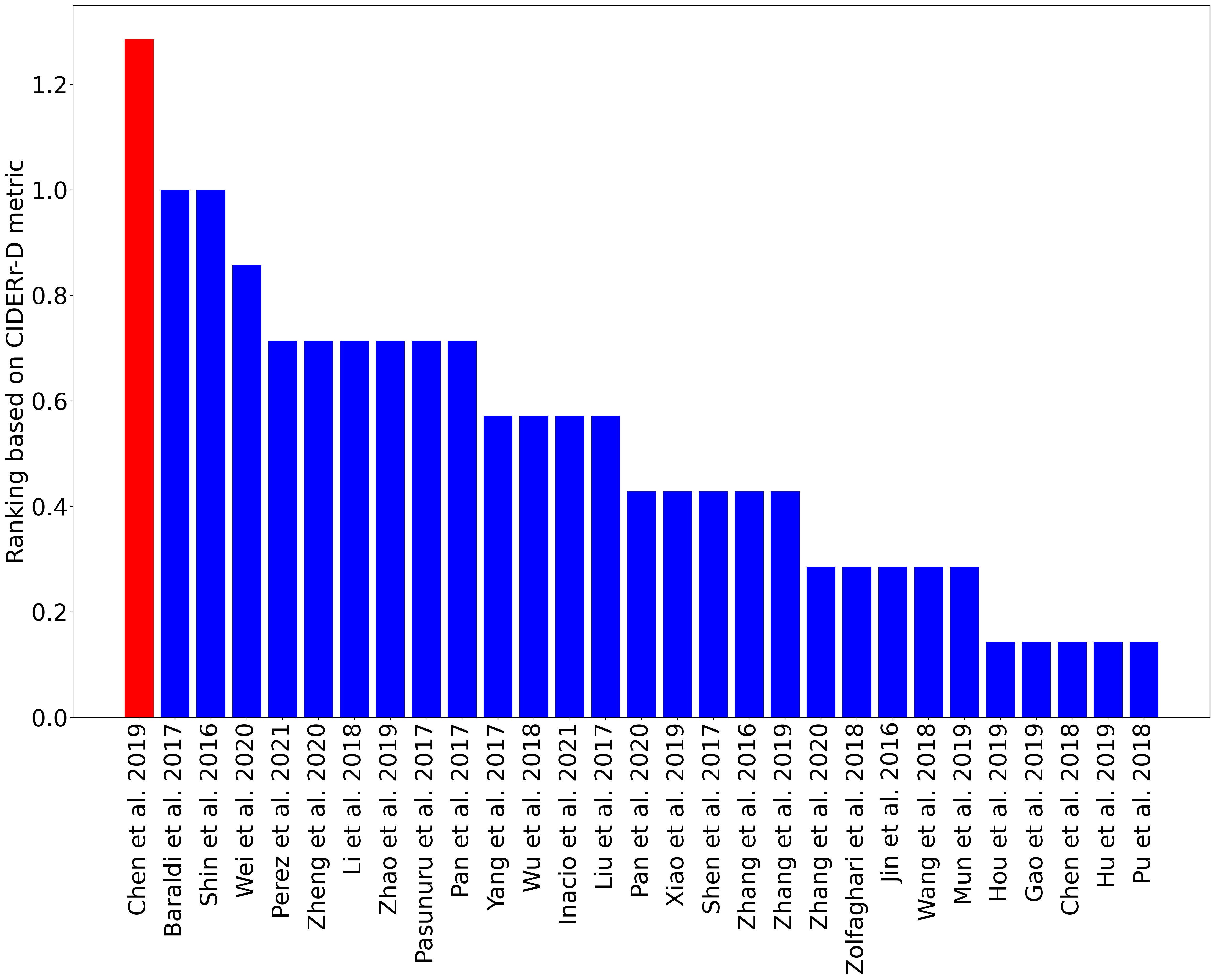} &    \includegraphics[scale=0.13]{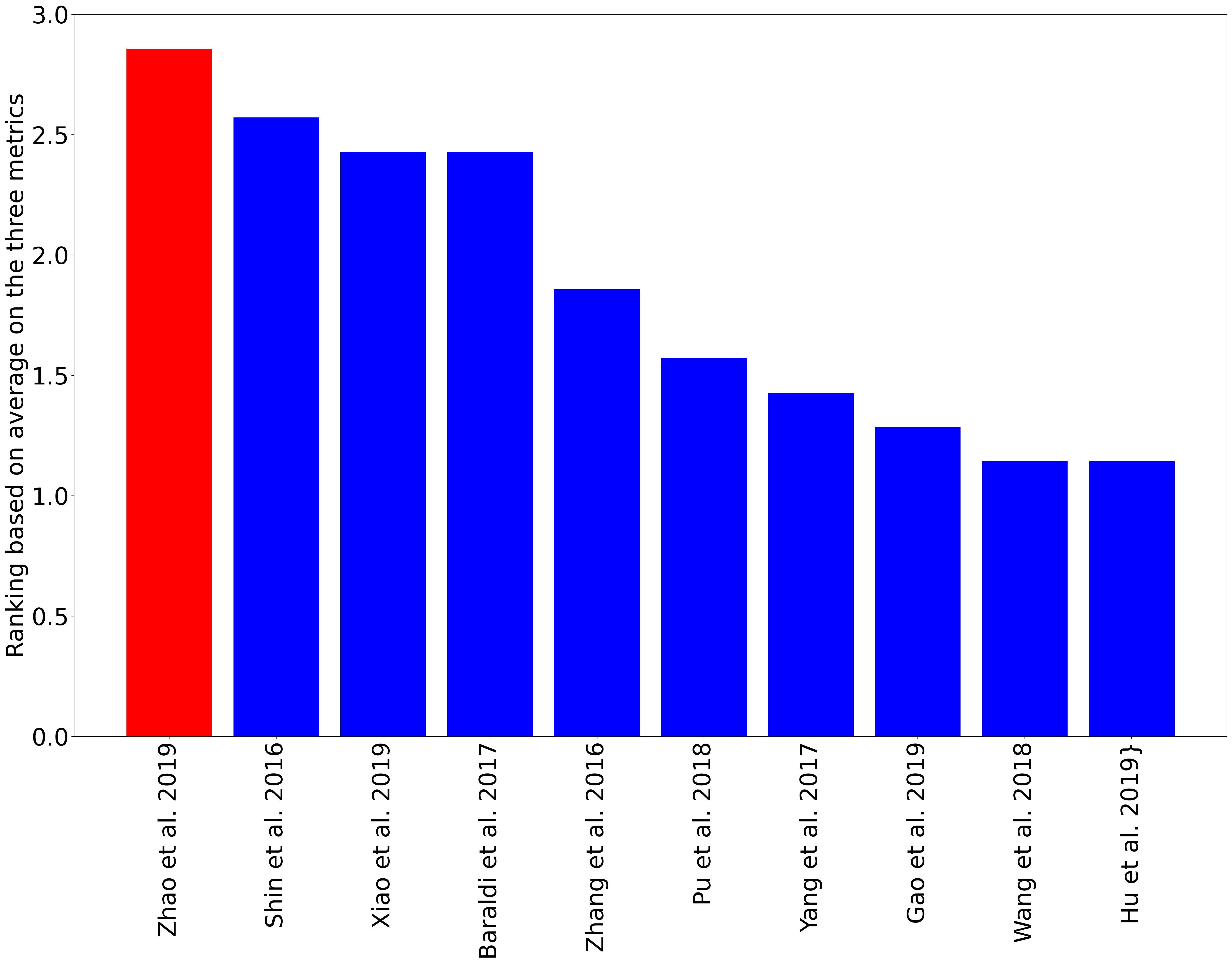} \\
			(c)  Ranking based on CIDERr-D & (d)  Ranking based on average of the 3 metrics \\[6pt]
		\end{tabular}
		\caption{Results based on METEOR metrics in most used datasets}
		\label{fig:rankings}
	\end{figure*}

	These ranking results can be seen in Figure~\ref{fig:rankings}(a), (b), and (c), respectively. The best work over all the datasets with the METEOR metric was the proposed by~\cite{zhao2019cam} (marked in red); the higher the bar value the better the ranking is, so those works with lower bar height have the worst performance.
	Similar behavior from BLEU-4, in which two works reported best results, the proposed in~\cite{xiao2019video}, and in~\cite{chen2019generating}.
	And, using CIDERr-D, the best method ranked is the proposed by~\cite{chen2019generating}

	Since not all the works utilized all the datasets to test their solutions, it is not straightforward to rank results over all the datasets; hence, a new ranking was calculated, joining the prior three ones through the average of its previous values, and then sorted them to highest to lowest ones.
	That is, in the case of one paper having first place in one dataset but also second place in another dataset, it will have a better ranking than others only citing one dataset.
	The new ranking considers the result in each dataset and how many datasets the work used. The final ranking is shown in Figure~\ref{fig:rankings} (d) from these computations. With the above, a unique winner could be selected; in this case, the best result considering all the metrics and datasets is achieved by the method proposed by Xiong et al. (see~\cite{xiong2018move}). It is important to note that it is easy to observe the best results with the ranking.

	\subsection{About approaches}
	\label{sec:aboutapproaches}
	Once the best method in the literature review is determined, it is also essential to know what approaches are prevalent.

	The revision extracted the reported results for each dataset, the methods employed, and the follow-up of the references cited or compared. In this way, we state that most of the reported results related to video captioning are considered in this study.
	The first analysis identifies the technique used in the encoder-decoder framework due to most of the papers utilizing similar methods adding or proposing an innovation to achieve better performance.
	The convolutional neural networks (CNN) are present in most of the works; the second most employed network is the C3D, and in the last years, the R-CNN architecture has been increasingly used due to the capability to detect objects. Furthermore, the process of audio has been presented through the MFCC technique \cite{memon2009using,sahidullah2009use}.

\begin{figure}[!h]
		\centering
		{\includegraphics[width=.4\textwidth]{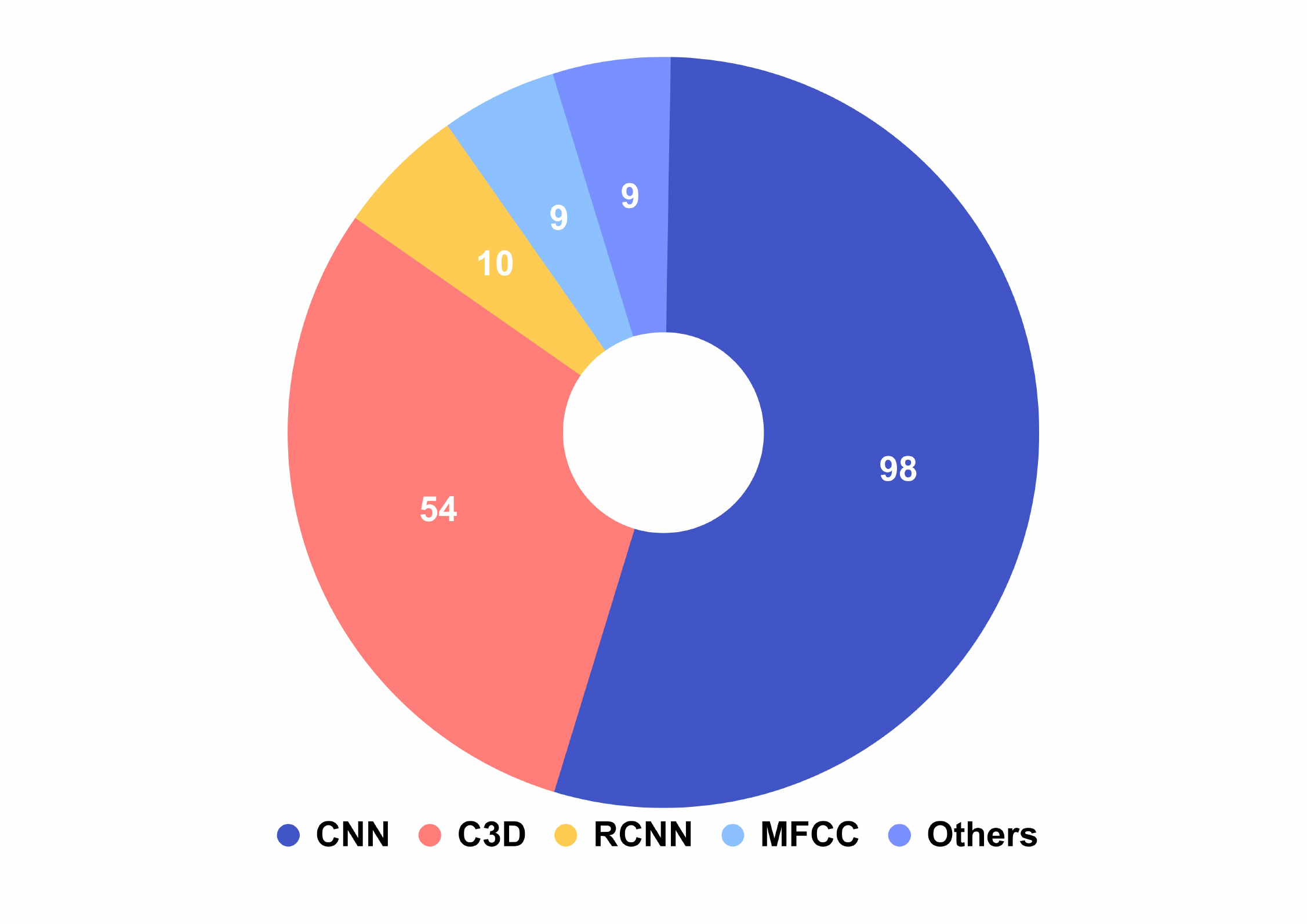}}
		\caption{Encoders used over 105 papers revised}
		\label{fig:encoder}
	\end{figure}

	Other methods have been used, and it is crucial to highlight that the total numbers presented here do not correspond to the 105 papers; the main reason is the fusion of features. Figure \ref{fig:encoder} shows a graphic distribution of approaches utilized to extract features in the decoder for video captioning. It is indispensable to mention that a few papers fusion different techniques for this purpose.

	The approaches for a decoder in video captioning are particularly emphasized in employing LSTM architectures, and this technique is a specific arrangement of a recurrent neural network (RNN).

\begin{figure}[!h]
		\centering
		{\includegraphics[width=.4\textwidth]{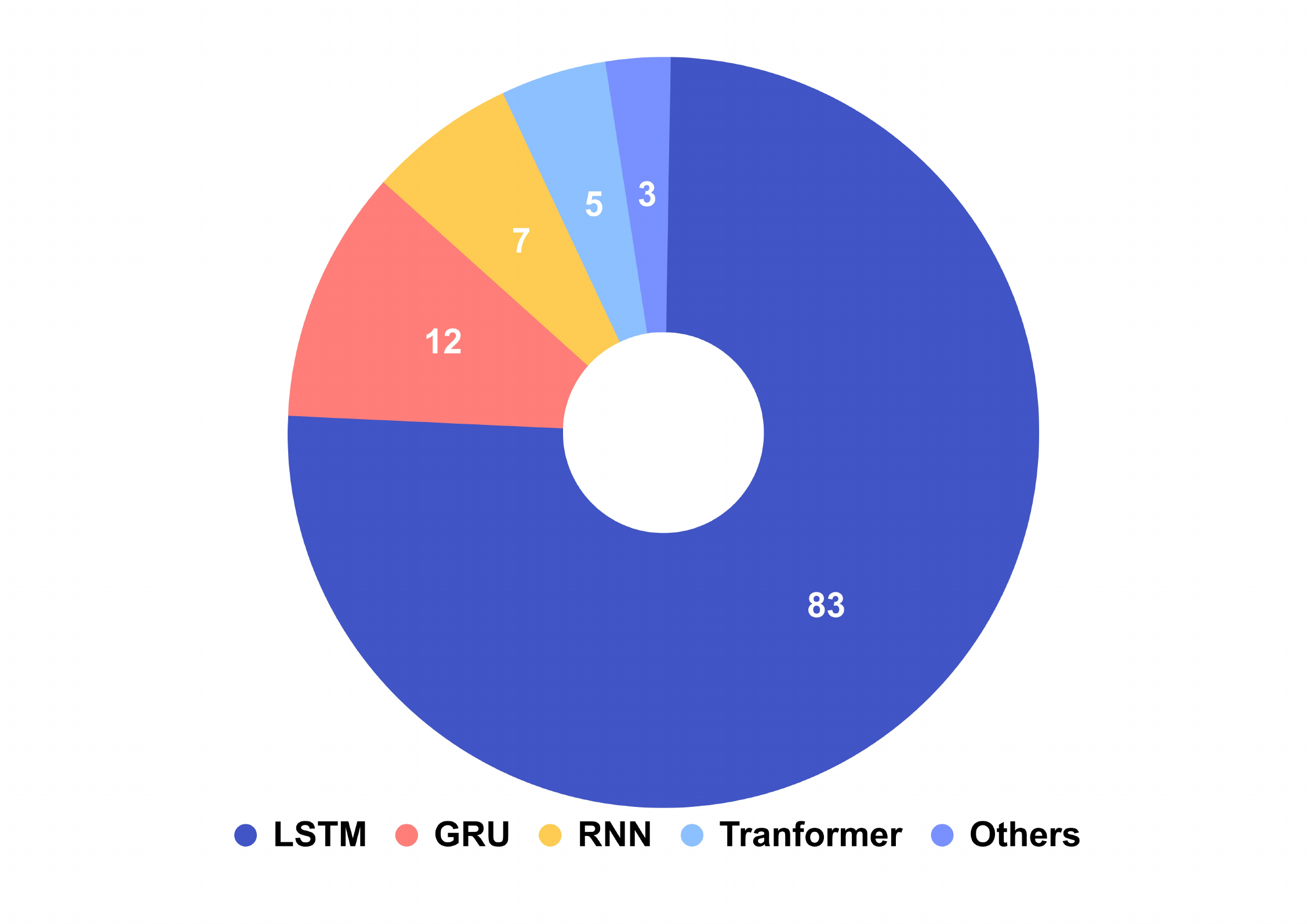}}
		\caption{Decoders used over 105 papers revised}
		\label{fig:decoder}
	\end{figure}

	However, GRUs and transformers have arisen in recent years; other methods have emerged but do not have much presence yet. Figure \ref{fig:decoder} displays a visual distribution of practices employed to covert the features of the decoder in captions.

\begin{figure}[!h]
		\centering
		\includegraphics[scale=0.3]{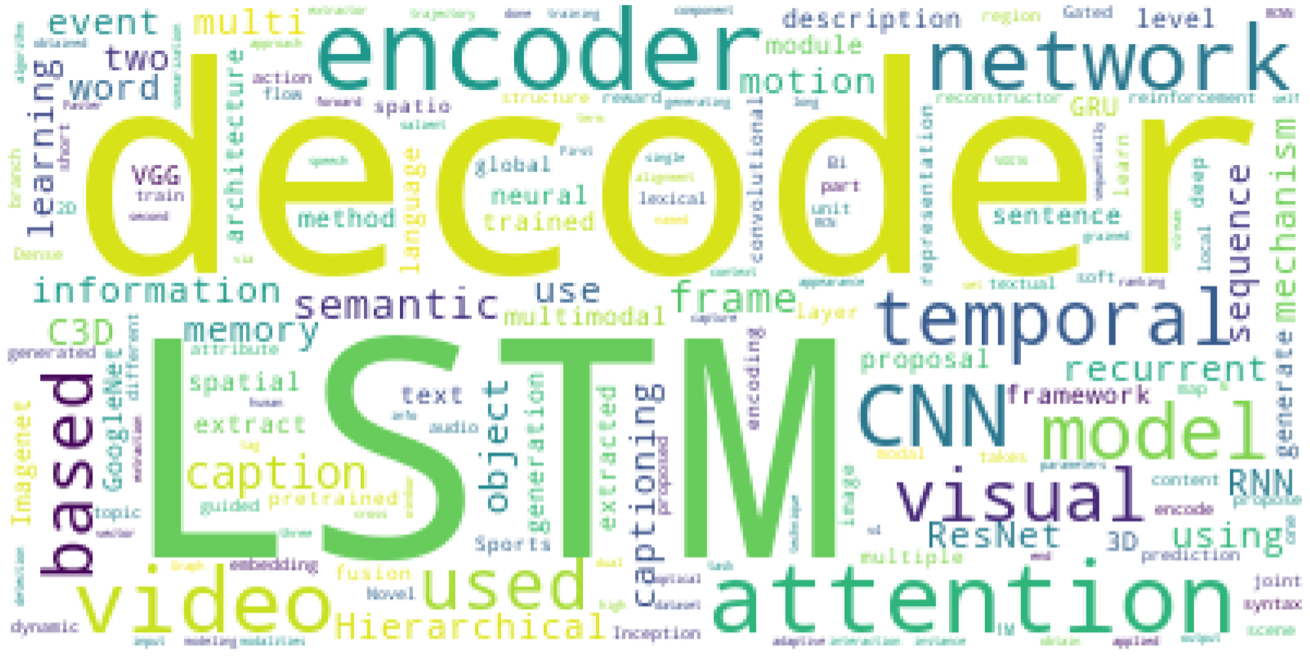}
		\caption{Wordcloud from the used methods or approaches}
		\label{fig:wordcloudmethods}
	\end{figure}

	To visualize the most frequent methods employed in the reviewed works, Figure~\ref{fig:wordcloudmethods} shows in a word cloud the most frequent words which correspond to decoder, LSTM, encoder, CNN, visual attention, RNN, and semantic, among others.

\begin{figure}[!h]
		\centering
		\includegraphics[scale=0.3]{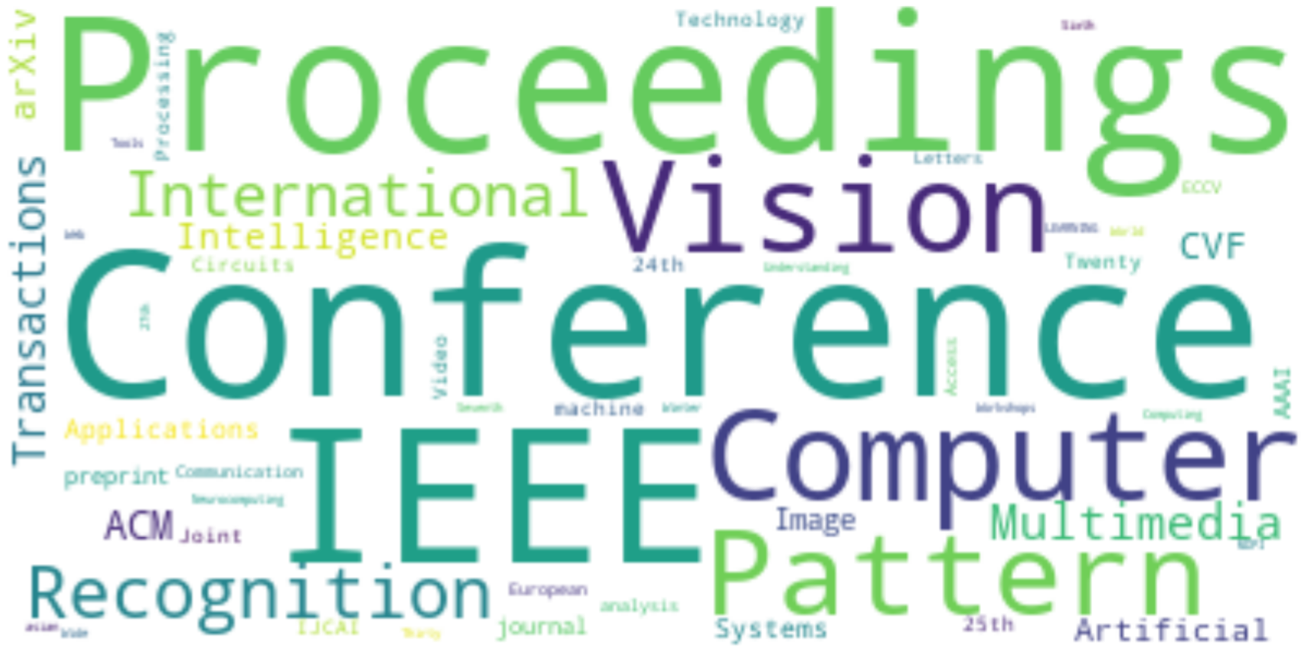}
		\caption{Wordcloud of where were published the papers revised}
		\label{fig:wordcloudpublicaciones}
	\end{figure}

	Finally, we consider it interesting to analyze where the methods dealing with video captioning tasks have been published; for this reason, Figure~\ref{fig:wordcloudpublicaciones} shows a word cloud with the most frequent words related to the name of publication, either a journal, conference, or others. Here, we can observe that conferences and proceedings are the most frequent because most revised works were published in high-quality conferences.

	As calculated before, the best three overall works are ~\cite{zhao2019cam}, ~\cite{shin2016beyond}, and \cite{xiao2019video}, first, second, and third place, respectively.

	The work proposed by  Zhao et al. in~\cite{zhao2019cam}, tries to exploit the correlation for both images and text features because sometimes there is a large non-correlation between them. To do that, Zhao et al. propose a co-attention model bases on a classical RNN network, named CAM-RNN.
	The CAM comprises three parts; visual attention uses region level and frame level attention to extract visual features from more correlated visual features (1).
	Also, the text attention layer deal with more correlated text features based on previous ones (2). Moreover, the balancing gate (3) regulates the influence of the visual features in an adaptively way. In the end, the RNN does the decoder task.
	One of the main differences between this work to the other ones, is the use in a joint way, of the visual and text features based on correlation, along with the weighing of the influence of visual features in the text generation. Another outstanding point is that Zhao et al. use four of the most used datasets to do their experiments, giving it more relevance to our ranking strategy.
	For the feature extractor, the CNN in some architectures such as VGGNet16, GoogLeNet (the first two pre-trained on ImageNet), and C3D were used (pre-trained on Sports-1M).
	A curious aspect of the experiments conducted by Zhao et al., is that only videos with total frames of 160 were used; in the case of longer ones, their frames were sampled uniformly to satisfy this condition, and shorter ones were padded with zeros.

	The second-best approach was proposed by Shin et al.~\cite{shin2016beyond}, who proposed a temporal segmentation strategy based on action localization from the videos. To extract motion features, they used the Dense Trajectories approach~\cite{wang2013action}.
	They also used a temporal sliding window and non-maximum suppression for temporal action localization; these temporal sliding windows were fixed in specific lengths of 30, 60, 90, and 120 frames. Furthermore, Shin et al. used backward co-reference resolution to caption generation from multiples frames, which means each segment of the video generates a caption, but at the end, a unique caption must describe the entire video content. To deal with that, the authors used first a gender annotator from Stanford CoreNLP~\cite{manning2014stanford}, which assigns likely gender to tokens, then the co-reference resolution was employed with POS (Part of Speech) tagger and lemmatization strategies. Another issue to deal with is the appropriate transition in captions with connective words (e.g., ``then'' word); to solve that, they ran a CKY parser for PCFG trained on Penn Treebank~\cite{marcinkiewicz1994building}. Later, the instances, not considering the connective words, are computed as 300-dimensional vectors using the Sentence2Vec~\cite{le2014distributed} approach. Ultimately, to do the caption generation, they perform matching of vectorization using the L2 distance.
	The most outstanding feature of the proposal of Shin et al. is to exploit the content of the video by segments of different sizes, to look for action, not classifying them, but localizing them. The above involves a lot of specific and cumbersome (at least we think) tasks to training specifically. One may think, if there is no action in the video, what will be the result? In that case, they only employ the middle frame to generate the video caption.
	As well as the solution proposed in the first place (see \cite{zhao2019cam}), Shin et al. tested their approach with three different datasets.

	Finally, the third-place approach by Xiao et al.~\cite{xiao2019video} employs a reinforced adaptive attention model (RAAM). In this case, they used CNN and Bi-LSTM networks to generate the video features. They try to avoid attention to each word.
	Furthermore,  Xiao et al. proposed a mixed loss training using the policy gradient not only to minimize the word-level cross-entropy, but also the sentence level.
	Their main contributions lie in the three unidirectional LSTM layers for the decoding task. As well as their combination of visual and contextual information regarding the next-word generation. In this case, Xiao et al. employed the two most used datasets for training, MSVD, and MPII.
	In general, the improvements of the architecture proposed by Xiao et al. are concrete; we think they take care of more detailed aspects than the first and second-best approaches analyzed in this review.

	\section{Areas to be improved}
	\label{sec:areastobeimproved}
	There are some opportunity areas that could be attended. Some datasets that have been tested achieving low results could be the source of the lack of generalization of the proposed methods to deal with this task.
	Some video content features are sometimes missed or non-used, especially those related to speech recognition and movement cues in general.

	The lack of exploitation of prior upstanding works related to video action recognition, video motions, trajectories analysis, object, and classification is poor.
	We saw that best works take advantage of many motion features related to time or temporal cues.
	Another essential aspect that could be improved is the evaluation metrics used to report performance (and therefore rank the models), because the most used are those not related to semantic aspect; we are aware of new metrics such as BLEURT~\cite{sellam-etal-2020-bleurt}, BERTScore~\cite{zhang2019bertscore}, BERTr~\cite{mathur2019putting}, CLIPScore~\cite{hessel2021clipscore} which use semantic or contextual embeddings to evaluate the semantic relationship of words.
	These approaches take advantage of considering the whole sentence for their embeddings ({\it e.g.}, BERTScore's bidirectionality) being able to distinguish between polysemous words and homonyms, due to the context of the description (or references in the training phase).

	As it had been exposed, the first generation, such as the BLEU metric, relies on measuring the superficial similarity between the sentences, like lexical or grammatical matching instead of semantics. In this case, none of the 105 revised papers used semantically related metrics. All the reviewed publications evaluate their approaches based on the classical first-generation (n-grams based) metrics.
	Nowadays, although BLEU or CIDERr-D metrics are still commonly used on tasks such as Image Captioning, the new generation of metrics is starting to have more relevance, but in Video Captioning it is not the case yet.

	Through all the papers reviewed, there are not many application problems. Most of the works consulted only apply their methods to the publicly available datasets leaving their work on a basic-science level. We find a huge opportunity to take these models into real-life scenarios to assess their performance in such tasks. We think only reporting performance in terms of known standard metrics is the first step, but further efforts must be done to apply such models to actual practical implementations.

	Among the few works that go a step beyond, the video captioning task is mainly applied to the description or synopsis of films to help people with visual impairment; or for the description of recipes. Its use has also been extended to narrate actions in sports, but there are few related works mentioned. Moreover, other areas have been barely explored, for example, smart video surveillance. This application would be helpful in the security sector since a natural language report could be generated on outstanding events or unusual actions.

	We acknowledge that state-of-the-art is still far from the point where there would be a {\it one-size-fits-all} solution for different video captioning tasks. For instance, the description of videos with lots of information in a short time-lapse implies one type of challenge, since there are more sequences to process and detect changes and generate enough captions from it, for example, in sports narration. On the other hand, in surveillance videos that have a longer duration, the scenes to describe can be comparatively shorter; but the challenge is to detect the relevant events to be captioned and do so in real-time, to take proper immediate action.

	We can also mention automatic subtitling of videos, scenes, or camera footage for visually impaired people, where the need for real-time performance meets the requirements of extracting meaningful information while discarding not relevant actions. If we take this a step forward, aiding people to navigate cumbersome environments demands the models' safety and robustness, maximizing true-positive rates for risk detection and minimizing the false-positive rates for user comfort.

	We do not find any glance of tackling the needs mentioned above at this moment and on the reviewed papers. Hence, we believe huge improvements should be sought in this direction to understand each approach's range of applications truly. We leave this review as an academic effort to gain insight into tracing an overview of the video captioning task, but we would like to contribute to the future by taking these models and research to real-life environments to understand their performance better.

	\section{Conclusions}
	\label{sec:conclusions}
	In this manuscript was presented a comparative review with a revision from the years 2016 to 2021. In this literature revision, more than 105 papers were analyzed and 105 works were used for a comparison between them. Also, most used datasets and metrics are evaluated and described.
	Through this period, we compared all the works based on different metrics, namely, METEOR, BLEU-4, and CIDERr-D. They were also compared taking into account the reported dataset used to evaluate the proposed methods.
	To obtain a winner over this analyzed period, we generate an easy ranking scheme where those works achieving the best results on each dataset got high scores, and with more datasets used in their experiments, the score they get is higher.
	We calculate, based on this ranking procedure, the best three methods, which are the ones reported by Zhao et al.~\cite{zhao2019cam}, Shin et al.~\cite{shin2016beyond}, and Xiao et al.\cite{xiao2019video}, first, second, and third place, respectively.
	Finally, after the complete analysis, some insights and improvement opportunities are mentioned.

	\section*{CRediT authorship contribution statement}
	\textbf{Daniela Moctezuma:} Conceptualization, Methodology, Data Curation, Software, Investigation, Writing – original draft, Visualization, Funding acquisition. \textbf{Tania Ramírez-delReal:} Conceptualization, Methodology, Data Curation, Investigation, Writing – original draft, Visualization. \textbf{Guillermo Ruiz:} Writing – review \& editing, Visualization, Validation. \textbf{Othón González:} Visualization, Writing – review \& editing, Visualization, Validation.

	\section*{Declaration of competing interest}
	The authors declare that they have no known competing financial interests or personal relationships that could have appeared to influence the work reported in this paper.

	\section*{Acknowledgments}

	This work has been done through CONACYT (National Council of Science and Technology from Mexico) support with Ciencia Básica grant with project ID A1-S-34811.

	\section*{Appendix}
	\label{sec:appendix}

	Tables show all the papers reviewed organized by dataset.

	\begin{table*}[h]
		\resizebox{18cm}{!}
		{
			\begin{tabular}{lrrrrrrrrrrrr}
				\toprule
				Work &  BLEU-1 &  BLEU-2 &  BLEU-3 &  BLEU-4 &  METEOR &  CIDERr-D &  ROUGE-L &  Average-Recall (AR) &  SPICE &  FCE &  RE &  Self-BLEU \\
				\midrule
				\cite{shin2016beyond} &    - &    - &    - &   6.30 &   10.40 &     17.70 &      - &                  - &    - &  - & - &        - \\
				\cite{pan2016hierarchical} &  81.10 &  68.60 &  57.80 &  46.70 &   33.90 &       - &      - &                  - &    - &  - & - &        - \\
				\cite{yu2016video} &  81.50 &  70.40 &  60.50 &  49.90 &   32.60 &     65.80 &      - &                  - &    - &  - & - &        - \\
				\cite{zanfir2016spatio} &  81.50 &  70.80 &  61.50 &  50.60 &   32.40 &       - &      - &                  - &    - &  - & - &        - \\
				\cite{fakoor2016memory} &  79.40 &  67.10 &  56.80 &  46.10 &   31.80 &     62.70 &      - &                  - &    - &  - & - &        - \\
				\cite{pan2016jointly} &  78.80 &  66.00 &  55.40 &  45.30 &   31.00 &       - &      - &                  - &    - &  - & - &        - \\
				\cite{venugopalan2016improving} &    - &    - &    - &  42.10 &   31.40 &       - &      - &                  - &    - &  - & - &        - \\
				\cite{ballas2015delving} &    - &    - &    - &  43.26 &   31.60 &     68.01 &      - &                  - &    - &  - & - &        - \\
				\cite{zhang2016automatic} &    - &    - &    - &    - &   31.10 &       - &      - &                  - &    - &  - & - &        - \\
				\cite{yang2017catching} &    - &    - &    - &  51.10 &   33.60 &     74.80 &      - &                  - &    - &  - & - &        - \\
				\cite{baraldi2017hierarchical} &    - &    - &    - &  42.50 &   32.40 &     63.50 &      - &                  - &    - &  - & - &        - \\
				\cite{xu2017learning} &  82.30 &  71.10 &  61.80 &  52.30 &   33.60 &     70.40 &      - &                  - &    - &  - & - &        - \\
				\cite{pasunuru2017multi} &    - &    - &    - &  54.50 &   36.00 &     92.40 &    72.80 &                  - &    - &  - & - &        - \\
				\cite{zhang2017task} &    - &    - &    - &  45.80 &   33.30 &     73.00 &    69.70 &                  - &    - &  - & - &        - \\
				\cite{gao2017video} &  81.80 &  70.80 &  61.10 &  50.80 &   33.30 &     74.80 &      - &                  - &    - &  - & - &        - \\
				\cite{liu2017video} &  80.20 &  69.00 &  60.10 &  51.10 &   32.60 &       - &      - &                  - &    - &  - & - &        - \\
				\cite{pan2017video} &    - &    - &    - &  54.50 &   36.00 &     92.40 &    72.80 &                  - &    - &  - & - &        - \\
				\cite{tu2017video} &  82.60 &  71.40 &  61.60 &  51.10 &   32.70 &     67.50 &      - &                  - &    - &  - & - &        - \\
				\cite{gan2017semantic} &    - &    - &    - &  51.10 &   33.50 &     77.70 &      - &                  - &    - &  - & - &        - \\
				\cite{jain2017recurrent} &    - &    - &    - &  45.70 &   31.90 &     57.30 &      - &                  - &    - &  - & - &        - \\
				\cite{dong2017improving} &    - &    - &    - &  44.60 &   29.70 &       - &      - &                  - &    - &  - & - &        - \\
				\cite{song2017hierarchical} &  82.90 &  72.20 &  63.00 &  53.00 &   33.60 &     73.80 &      - &                  - &    - &  - & - &        - \\
				\cite{liu2017hierarchical} &  78.00 &  65.20 &  54.90 &  44.30 &   32.10 &     68.40 &    68.90 &                  - &    - &  - & - &        - \\
				\cite{zhu2017bidirectional} &    - &    - &    - &  49.45 &   33.39 &     75.45 &      - &                  - &    - &  - & - &        - \\
				\cite{song2018deterministic} &  82.90 &  72.60 &  63.50 &  53.30 &   33.80 &     74.80 &      - &                  - &    - &  - & - &        - \\
				\cite{wang2018hierarchical} &    - &    - &    - &  52.90 &   33.80 &     74.50 &      - &                  - &    - &  - & - &        - \\
				\cite{zhang2019hierarchical} &  83.10 &  73.00 &  64.30 &  55.10 &   35.30 &     83.30 &      - &                  - &    - &  - & - &        - \\
				\cite{daskalakis2018learning} &  78.11 &  66.43 &  55.93 &  45.02 &   33.80 &     63.28 &    69.62 &                  - &    - &  - & - &        - \\
				\cite{chen2018less} &    - &    - &    - &  52.30 &   33.30 &     76.50 &    69.60 &                  - &    - &  - & - &        - \\
				\cite{wang2018m3} &  82.45 &  72.43 &  62.78 &  52.82 &   33.31 &       - &      - &                  - &    - &  - & - &        - \\
				\cite{wu2018interpretable} &    - &    - &    - &  51.70 &   34.00 &     74.90 &      - &                  - &    - &  - & - &        - \\
				\cite{long2018video} &  83.00 &  71.90 &  63.00 &  52.00 &   33.50 &     72.10 &      - &                  - &    - &  - & - &        - \\
				\cite{yang2018video} &    - &    - &    - &  42.90 &   30.40 &       - &      - &                  - &    - &  - & - &        - \\
				\cite{wang2018reconstruction} &    - &    - &    - &  52.30 &   34.10 &     80.30 &    69.80 &                  - &    - &  - & - &        - \\
				\cite{li2018multimodal} &    - &    - &    - &  48.00 &   31.60 &     68.80 &      - &                  - &    - &  - & - &        - \\
				\cite{wu2018multi} &    - &    - &    - &  46.46 &   33.72 &     75.46 &      - &                  - &    - &  - & - &        - \\
				\cite{bin2018describing} &  79.00 &  60.50 &  48.40 &  37.30 &   30.30 &       - &      - &                  - &    - &  - & - &        - \\
				\cite{xu2018sequential} &    - &    - &    - &  51.00 &   35.15 &     86.04 &      - &                  - &    - &  - & - &        - \\
				\cite{pu2018adaptive} &    - &    - &    - &  54.27 &   38.03 &     78.31 &      - &                  - &    - &  - & - &        - \\
				\cite{yan2019stat} &    - &    - &    - &  52.00 &   33.33 &     73.80 &      - &                  - &    - &  - & - &        - \\
				\cite{wu2019convolutional} &    - &    - &    - &  54.10 &   35.15 &     82.75 &      - &                  - &    - &  - & - &        - \\
				\cite{xu2018dual} &    - &    - &    - &  53.00 &   34.70 &     79.40 &    65.90 &                  - &    - &  - & - &        - \\
				\cite{shi2019watch} &    - &    - &    - &  51.70 &   34.30 &     86.70 &    71.90 &                  - &    - &  - & - &        - \\
				\cite{xiao2019video} &  84.20 &  74.10 &  65.00 &  55.40 &   35.60 &     85.50 &    71.10 &                  - &    - &  - & - &        - \\
				\cite{aafaq2019spatio} &    - &    - &    - &  47.90 &   35.90 &     78.10 &    71.50 &                  - &    - &  - & - &        - \\
				\cite{tang2019rich} &  82.80 &  71.70 &  62.40 &  52.40 &   35.70 &     84.30 &    72.20 &                  - &    - &  - & - &        - \\
				\cite{zhang2019object} &    - &    - &    - &  56.90 &   36.20 &     90.60 &      - &                  - &    - &  - & - &        - \\
				\cite{xu2019multi} &  82.10 &  71.60 &  61.40 &  53.00 &   32.90 &     75.10 &    69.80 &                  - &    - &  - & - &        - \\
				\cite{chen2019motion} &    - &    - &    - &  53.40 &   35.00 &     86.70 &      - &                  - &    - &  - & - &        - \\
				\cite{pei2019memory} &    - &    - &    - &  48.60 &   35.10 &     92.20 &    71.90 &                  - &    - &  - & - &        - \\
				\cite{hou2019joint} &    - &    - &    - &  52.80 &   36.10 &     87.80 &    71.80 &                  - &    - &  - & - &        - \\
				\cite{hu2019hierarchical} &  86.80 &  75.00 &  65.10 &  54.70 &   35.20 &     91.30 &    72.50 &                  - &    - &  - & - &        - \\
				\cite{gao2019hierarchical} &  83.30 &  73.60 &  64.60 &  54.30 &   33.50 &     72.80 &      - &                  - &    - &  - & - &        - \\
				\cite{guo2019exploiting} &  83.80 &  73.80 &  64.50 &  54.50 &   34.50 &     79.30 &      - &                  - &    - &  - & - &        - \\
				\cite{li2019end} &    - &    - &    - &  50.30 &   34.10 &     87.50 &    70.80 &                  - &    - &  - & - &        - \\
				\cite{li2019residual} &  82.80 &  72.30 &  63.10 &  53.40 &   34.30 &     72.90 &      - &                  - &    - &  - & - &        - \\
				\cite{wang2019controllable} &    - &    - &    - &  42.00 &   28.20 &     48.70 &    61.60 &                  - &    - &  - & - &        - \\
				\cite{xiao2020video} &    - &    - &    - &  54.10 &   36.10 &     86.10 &    72.40 &                  - &    - &  - & - &        - \\
				\cite{shi2020video} &    - &    - &    - &  51.70 &   33.00 &     71.00 &      - &                  - &    - &  - & - &        - \\
				\cite{pan2020spatio} &    - &    - &    - &  52.20 &   36.90 &     93.00 &    73.90 &                  - &    - &  - & - &        - \\
				\cite{liu2020sibnet} &    - &    - &    - &  55.70 &   35.50 &     88.80 &    72.60 &                  - &    - &  - & - &        - \\
				\cite{zhang2019reconstruct} &    - &    - &    - &  52.30 &   34.10 &     80.30 &    69.80 &                  - &    - &  - & - &        - \\
				\cite{zhang2020object} &    - &    - &    - &  54.30 &   36.40 &     95.20 &    73.90 &                  - &    - &  - & - &        - \\
				\cite{lei2021video} &    - &    - &    - &  52.20 &   35.60 &     83.70 &    72.70 &                  - &    - &  - & - &        - \\
				\cite{chen2021motion} &    - &    - &    - &  55.80 &   36.90 &     74.50 &    98.50 &                  - &    - &  - & - &        - \\
				\cite{perez2021improving} &    - &    - &    - &  64.40 &   41.90 &    111.50 &    79.50 &                  - &    - &  - & - &        - \\
				\bottomrule
			\end{tabular}
		}
		\caption{All results on MSVD dataset}
		\label{apendice:allmsvd}
	\end{table*}

	\begin{table*}[h]
		\resizebox{18cm}{!}
		{
			\begin{tabular}{lrrrrrrrrrrrr}
				\toprule
				Work &  BLEU-1 &  BLEU-2 &  BLEU-3 &  BLEU-4 &  METEOR &  CIDERr-D &  ROUGE-L &  Average-Recall (AR) &  SPICE &  FCE &  RE &  Self-BLEU \\
				\midrule
				\cite{dong2016early} &    - &    - &    - &  38.70 &   26.90 &     45.90 &    58.70 &                  - &    - &  - & - &        - \\
				\cite{ramanishka2016multimodal}&    - &    - &    - &  40.70 &   28.60 &     46.50 &    61.00 &                  - &    - &  - & - &        - \\
				\cite{shetty2016frame}&    - &    - &    - &  41.11 &   27.70 &     46.40 &    59.60 &                  - &    - &  - & - &        - \\
				\cite{jin2016describing}&    - &    - &    - &  42.60 &   28.80 &     61.70 &    46.70 &                  - &    - &  - & - &        - \\
				\cite{nian2017learning} &    - &    - &    - &  41.70 &   28.90 &     51.40 &    62.10 &                  - &    - &  - & - &        - \\
				\cite{xu2017learning} &    - &    - &    - &  36.50 &   26.50 &     41.00 &    59.80 &                  - &    - &  - & - &        - \\
				\cite{pasunuru2017multi} &    - &    - &    - &  40.80 &   28.80 &     47.10 &    60.20 &                  - &    - &  - & - &        - \\
				\cite{zhang2017task} &    - &    - &    - &  35.50 &   28.20 &     42.70 &    59.10 &                  - &    - &  - & - &        - \\
				\cite{gao2017video} &    - &    - &    - &  38.00 &   26.10 &     43.20 &      - &                  - &    - &  - & - &        - \\
				\cite{chen2017video} &    - &    - &    - &  39.20 &   27.50 &     48.70 &    60.30 &                  - &    - &  - & - &        - \\
				\cite{shen2017weakly} &    - &    - &    - &  33.70 &   25.90 &     56.90 &    32.60 &                  - &    - &  - & - &        - \\
				\cite{tu2017video} &    - &    - &    - &  37.40 &   26.60 &     41.50 &      - &                  - &    - &  - & - &        - \\
				\cite{song2017hierarchical}&    - &    - &    - &  38.30 &   26.30 &       - &      - &                  - &    - &  - & - &        - \\
				\cite{liu2017hierarchical}&    - &    - &    - &  37.10 &   26.70 &     41.00 &    59.00 &                  - &    - &  - & - &        - \\
				\cite{phan2017consensus}&   82.8 &   69.5 &  56.20 &  44.10 &   29.10 &     49.70 &    62.40 &                  - &    - &  - & - &        - \\
				\cite{hori2017attention}&    - &    - &    - &  39.70 &   25.50 &     40.00 &      - &                  - &    - &  - & - &        - \\
				\cite{song2018deterministic} &    - &    - &    - &  39.80 &   26.10 &     40.90 &    59.30 &                  - &    - &  - & - &        - \\
				\cite{wang2018hierarchical} &    - &    - &    - &  39.90 &   28.30 &     40.90 &      - &                  - &    - &  - & - &        - \\
				\cite{chen2018less} &    - &    - &    - &  41.30 &   27.70 &     44.10 &    59.80 &                  - &    - &  - & - &        - \\
				\cite{wang2018m3} &   73.6 &   59.3 &  48.26 &  38.13 &   26.58 &       - &      - &                  - &    - &  - & - &        - \\
				\cite{long2018video} &    - &    - &    - &  41.30 &   28.70 &     48.00 &    61.70 &                  - &    - &  - & - &        - \\
				\cite{ wang2018video} &    - &    - &    - &  41.30 &   28.70 &     48.00 &    61.70 &                  - &    - &  - & - &        - \\
				\cite{yang2018video} &    - &    - &    - &  36.00 &   26.10 &       - &      - &                  - &    - &  - & - &        - \\
				\cite{ chen2018tvt} &    - &    - &    - &  42.46 &   28.24 &     48.53 &    61.07 &                  - &    - &  - & - &        - \\
				\cite{ wang2018reconstruction} &    - &    - &    - &  39.10 &   26.60 &     42.70 &    59.30 &                  - &    - &  - & - &        - \\
				\cite{ li2018multimodal} &   76.1 &   62.1 &  49.10 &  37.50 &   26.40 &       - &      - &                  - &    - &  - & - &        - \\
				\cite{wu2018multi} &    - &    - &    - &  38.10 &   27.20 &     42.10 &      - &                  - &    - &  - & - &        - \\
				\cite{bin2018describing} &   78.9 &   60.4 &  46.10 &  33.90 &   26.20 &       - &      - &                  - &    - &  - & - &        - \\
				\cite{pu2018adaptive}&    - &    - &    - &  45.01 &   29.98 &     51.41 &      - &                  - &    - &  - & - &        - \\
				\cite{yan2019stat}&    - &    - &    - &  39.30 &   27.10 &     43.80 &      - &                  - &    - &  - & - &        - \\
				\cite{wu2019convolutional}&    - &    - &    - &  38.10 &   27.20 &     42.10 &      - &                  - &    - &  - & - &        - \\
				\cite{xu2018dual} &    - &    - &    - &  42.30 &   29.40 &     46.10 &    62.30 &                  - &    - &  - & - &        - \\
				\cite{shi2019watch}&    - &    - &    - &  43.20 &   28.00 &     48.30 &    62.00 &                  - &    - &  - & - &        - \\
				\cite{qi2019sports}&    - &    - &    - &  36.70 &   25.90 &     33.90 &      - &                  - &    - &  - & - &        - \\
				\cite{aafaq2019spatio}&    - &    - &    - &  38.30 &   28.40 &     48.10 &    60.70 &                  - &    - &  - & - &        - \\
				\cite{tang2019rich}&   81.1 &   67.2 &  53.70 &  41.40 &   29.00 &     48.90 &    61.30 &                  - &    - &  - & - &        - \\
				\cite{zhang2019object}&    - &    - &    - &  41.40 &   28.20 &     46.90 &      - &                  - &    - &  - & - &        - \\
				\cite{xu2019multi}&    - &    - &    - &  40.80 &   27.50 &     45.40 &    60.70 &                  - &    - &  - & - &        - \\
				\cite{chen2019motion}&    - &    - &    - &  45.40 &   28.60 &     50.10 &      - &                  - &    - &  - & - &        - \\
				\cite{pei2019memory}&    - &    - &    - &  40.40 &   28.10 &     47.10 &    60.70 &                  - &    - &  - & - &        - \\
				\cite{hou2019joint}&    - &    - &    - &  42.30 &   29.70 &     49.10 &    62.80 &                  - &    - &  - & - &        - \\
				\cite{chen2019generating}&    - &    - &    - &  44.91 &   29.61 &     51.80 &    62.81 &                  - &   6.85 &  - & - &        - \\
				\cite{gao2019hierarchical} &   83.3 &   73.6 &  64.60 &  54.30 &   33.50 &     72.80 &      - &                  - &    - &  - & - &        - \\
				\cite{guo2019exploiting}&   77.6 &   64.0 &  51.30 &  39.90 &   27.10 &     43.80 &      - &                  - &    - &  - & - &        - \\
				\cite{li2019end}&    - &    - &    - &  40.40 &   27.90 &     48.30 &    61.00 &                  - &    - &  - & - &        - \\
				\cite{lee2019deep}&    - &    - &    - &  36.60 &   23.80 &     27.10 &    52.40 &                  - &    - &  - & - &        - \\
				\cite{zhao2019cam} &    - &    - &    - &  36.20 &   27.90 &     38.80 &    58.80 &                  - &    - &  - & - &        - \\
				\cite{li2019residual} &   77.1 &   62.1 &  48.70 &  37.00 &   26.90 &     40.70 &      - &                  - &    - &  - & - &        - \\
				\cite{wang2019controllable}&    - &    - &    - &  41.70 &   27.90 &     48.40 &    61.00 &                  - &    - &  - & - &        - \\
				\cite{xiao2020video}&    - &    - &    - &  44.60 &   28.70 &     48.60 &    62.20 &                  - &    - &  - & - &        - \\
				\cite{shi2020video}&    - &    - &    - &  41.80 &   26.70 &     45.30 &      - &                  - &    - &  - & - &        - \\
				\cite{pan2020spatio}&    - &    - &    - &  40.50 &   28.30 &     47.10 &    60.90 &                  - &    - &  - & - &        - \\
				\cite{liu2020sibnet}&    - &    - &    - &  41.20 &   27.80 &     48.60 &    60.80 &                  - &    - &  - & - &        - \\
				\cite{zhang2019reconstruct}&    - &    - &    - &  39.20 &   27.50 &     48.70 &    60.30 &                  - &    - &  - & - &        - \\
				\cite{zhang2020object}&    - &    - &    - &  43.60 &   28.80 &     50.90 &    62.10 &                  - &    - &  - & - &        - \\
				\cite{wei2020exploiting}&    - &    - &    - &  38.50 &   26.90 &     43.70 &      - &                  - &    - &  - & - &        - \\
				\cite{yuan2020controllable}&    - &    - &    - &   5.01 &   16.25 &     12.22 &    30.41 &                  - &    - &  - & - &        - \\
				\cite{lei2021video}&    - &    - &    - &  41.30 &   28.20 &     48.60 &    61.90 &                  - &    - &  - & - &        - \\
				\cite{chen2021motion}&    - &    - &    - &  41.70 &   28.90 &     51.40 &    62.10 &                  - &    - &  - & - &        - \\
				\cite{perez2021improving}&    - &    - &    - &  46.40 &   30.40 &     51.90 &    64.70 &                  - &    - &  - & - &        - \\
				\cite{ji2021multi}&    - &    - &    - &  39.30 &   27.10 &       - &    59.50 &                  - &    - &  - & - &        - \\
				\bottomrule
			\end{tabular}
		}
		\caption{All results on MSR-VTT dataset}
		\label{apendice:allmsrvtt}
	\end{table*}

	\begin{table*}[h]
		\resizebox{18cm}{!}
		{
			\begin{tabular}{lrrrrrrrrrrrr}
				\toprule
				Work &  BLEU-1 &  BLEU-2 &  BLEU-3 &  BLEU-4 &  METEOR &  CIDERr-D &  ROUGE-L &  Average-Recall (AR) &  SPICE &  FCE &    RE &  Self-BLEU \\
				\midrule
				\cite{inacio2021osvidcap} &    - &    - &    - &  4.320 &    9.98 &     30.50 &    21.33 &                  - &    - &  - &   - &        - \\
				\cite{yu2021accelerated} &    - &    - &    - &    - &    5.82 &     10.87 &      - &                  - &    - &  - &   - &        - \\
				\cite{iashin2020multi} &    - &    - &   5.83 &  2.860 &   11.72 &       - &      - &                  - &    - &  - &   - &        - \\
				\cite{yuan2020controllable} &    - &    - &    - &  0.530 &    5.11 &      9.31 &    11.11 &                  - &    - &  - &   - &        - \\
				\cite{suin2020efficient} &    - &    - &   2.87 &  1.350 &    6.21 &     13.82 &      - &                53.40 &    - &  - &   - &        - \\
				\cite{wang2018bidirectional} &  18.99 &   8.84 &   4.41 &  2.300 &    9.60 &     12.68 &    19.10 &                  - &    - &  - &   - &        - \\
				\cite{mun2019streamlined} &  28.02 &  12.05 &   4.41 &  1.280 &   13.07 &     43.48 &      - &                  - &    - &  - &   - &        - \\
				\cite{zhou2018end} &    - &    - &   4.76 &  2.230 &    9.56 &       - &      - &                52.95 &    - &  - &   - &        - \\
				\cite{qi2019sports} &  26.60 &  13.90 &   8.20 &  4.900 &    9.90 &     24.60 &      - &                  - &    - &  - &   - &        - \\
				\cite{zhang2019show} &  22.76 &  10.12 &   4.26 &  1.640 &   10.71 &     31.41 &    22.85 &                  - &    - &  - &   - &        - \\
				\cite{hou2019joint} &    - &    - &    - &  1.900 &   11.30 &     44.20 &    22.40 &                  - &    - &  - &   - &        - \\
				\cite{li2018jointly} &  19.57 &   9.90 &   4.55 &  1.622 &   10.33 &     25.24 &      - &                  - &    - &  - &   - &        - \\
				\cite{xiong2018move} &  39.11 &  22.26 &  13.52 &  8.450 &   14.75 &     14.15 &    14.75 &                  - &    - &  - & 17.59 &       45.8 \\
				\cite{krishna2017dense} &  26.45 &  13.48 &   7.12 &  3.980 &    9.46 &     24.56 &      - &                  - &    - &  - &   - &        - \\
				\cite{zhang2019reconstruct} &    - &    - &    - &  1.740 &   10.47 &     38.43 &    23.49 &                  - &    - &  - &   - &        - \\
				\bottomrule
			\end{tabular}
		}
		\caption{All results on ActivityNet Captions dataset}
		\label{apendice:allactivity}
	\end{table*}

	\begin{table*}[h]
		\resizebox{18cm}{!}
		{
			\begin{tabular}{lrrrrrrrrrrrr}
				\toprule
				Work &  BLEU-1 &  BLEU-2 &  BLEU-3 &  BLEU-4 &  METEOR &  CIDERr-D &  ROUGE-L &  Average-Recall (AR) &  SPICE &  FCE &  RE &  Self-BLEU \\
				\midrule
				\cite{shin2016beyond} &    - &    - &    - &   0.40 &    4.70 &      8.90 &      - &                  - &    - &  - & - &        - \\
				\cite{pan2016hierarchical} &    - &    - &    - &    - &    6.80 &       - &      - &                  - &    - &  - & - &        - \\
				\cite{pan2016jointly} &    - &    - &    - &    - &    6.70 &       - &      - &                  - &    - &  - & - &        - \\
				\cite{venugopalan2016improving} &    - &    - &    - &    - &    6.70 &       - &      - &                  - &    - &  - & - &        - \\
				\cite{zhang2016automatic} &    - &    - &    - &    - &    6.70 &       - &      - &                  - &    - &  - & - &        - \\
				\cite{yang2017catching} &    - &    - &    - &    - &    6.90 &       - &      - &                  - &    - &  - & - &        - \\
				\cite{baraldi2017hierarchical} &    - &    - &    - &    - &    7.30 &       - &      - &                  - &    - &  - & - &        - \\
				\cite{ nian2017learning} &    - &    - &    - &    - &    5.70 &       - &      - &                  - &    - &  - & - &        - \\
				\cite{ pasunuru2017multi} &    - &    - &    - &    - &    7.40 &       - &      - &                  - &    - &  - & - &        - \\
				\cite{ liu2017video} &    - &    - &    - &    - &    6.90 &       - &      - &                  - &    - &  - & - &        - \\
				\cite{pan2017video} &    - &    - &    - &    - &    7.40 &       - &      - &                  - &    - &  - & - &        - \\
				\cite{yang2018video} &    - &    - &    - &    - &    6.30 &       - &      - &                  - &    - &  - & - &        - \\
				\cite{xu2018sequential} &    - &    - &    - &    - &    7.20 &       - &      - &                  - &    - &  - & - &        - \\
				\cite{pu2018adaptive} &    - &    - &    - &   2.08 &    7.12 &      9.14 &      - &                  - &    - &  - & - &        - \\
				\bottomrule
			\end{tabular}
		}
		\caption{All results on MVAD dataset}
		\label{apendice:allmvad}
	\end{table*}

	\begin{table*}[h]
		\resizebox{18cm}{!}
		{
			\begin{tabular}{lrrrrrrrrrrrr}
				\toprule
				Work &  BLEU-1 &  BLEU-2 &  BLEU-3 &  BLEU-4 &  METEOR &  CIDERr-D &  ROUGE-L &  Average-Recall (AR) &  SPICE &  FCE &  RE &  Self-BLEU \\
				\midrule
				\cite{ shin2016beyond} &    - &    - &    - &   13.0 &     4.8 &       7.5 &      - &                  - &    - &  - & - &        - \\
				\cite{ baraldi2017hierarchical} &    - &    - &    - &    1.0 &     7.0 &      10.8 &     16.7 &                  - &    - &  - & - &        - \\
				\cite{ nian2017learning} &    - &    - &    - &    - &     6.6 &       - &      - &                  - &    - &  - & - &        - \\
				\cite{pan2017video} &    - &    - &    - &   40.8 &    28.8 &      47.1 &     60.2 &                  - &    - &  - & - &        - \\
				\cite{xu2018dual} &    - &    - &    - &    1.9 &     7.9 &       - &      - &                  - &    - &  - & - &        - \\
				\cite{xiao2019video} &    - &    - &    - &    0.9 &     6.8 &      10.0 &      - &                  - &    - &  - & - &        - \\
				\cite{xu2019multi} &    - &    - &    - &    0.8 &     7.7 &       - &      - &                  - &    - &  - & - &        - \\
				\cite{ zhao2019cam} &    - &    - &    - &    - &     7.8 &       - &      - &                  - &    - &  - & - &        - \\
				\cite{yang2018video} &    - &    - &    - &    - &     7.2 &       - &      - &                  - &    - &  - & - &        - \\
				\cite{liu2017hierarchical} &   16.9 &    5.4 &    1.6 &    0.6 &     7.1 &       8.9 &     17.0 &                  - &    - &  - & - &        - \\
				\cite{pan2016jointly} &    - &    - &    - &    - &     7.3 &       - &      - &                  - &    - &  - & - &        - \\
				\cite{venugopalan2016improving} &    - &    - &    - &    - &     6.8 &       - &      - &                  - &    - &  - & - &        - \\
				\cite{zhang2016automatic} &    - &    - &    - &    - &     7.0 &       - &      - &                  - &    - &  - & - &        - \\
				\bottomrule
			\end{tabular}
		}
		\caption{All results on MPII dataset}
		\label{apendice:allmpii}
	\end{table*}

	\begin{table*}[h]
		\resizebox{18cm}{!}
		{
			\begin{tabular}{lrrrrrrrrrrrr}
				\toprule
				Work &  BLEU-1 &  BLEU-2 &  BLEU-3 &  BLEU-4 &  METEOR &  CIDERr-D &  ROUGE-L &  Average-Recall (AR) &  SPICE &  FCE &  RE &  Self-BLEU \\
				\midrule
				\cite{nian2017learning} & -    & -    & -    &  40.10 &   29.90 &     51.10 & -      & -                   & -    & -  & - & -        \\
				\cite{ li2017mam} &   80.1 &   66.1 &   54.7 &  41.30 &   32.20 &     53.90 &    68.80 &                  - &    - &  - & - & -        \\
				\cite{chen2017video} & -    & -    & -    &  47.56 &   34.21 &     79.57 &    70.45 & -                  & -    & -  & - & -        \\
				\cite{hori2017attention} & -    &    - & -    &  56.80 &   34.30 &     72.40 & -      &  -                 & -    &-   & - & -        \\
				\cite{zhao2018video} &   77.6 &   67.1 &   55.4 &  43.80 &   32.60 &     52.20 &    69.30 &    -               &  -   & -  & - & -        \\
				\cite{chen2018tvt} & -    & -    & -    &  53.21 &   35.23 &     86.76 & -      &    -               &  -   & -  & - & -        \\
				\cite{zolfaghari2018eco} & -    & -    &   62.6 &  53.50 &   35.00 &     85.80 &   -    &      -             & -    & -  & - & -        \\
				\cite{qi2019sports} &   81.2 &   69.7 &   61.3 &  50.90 &   33.50 &     70.30 &      - &    -               &    - & -  & - & -        \\
				\cite{chen2019generating} &  -   &    - &     -&  49.26 &   33.91 &     83.20 &    70.95 &                 -  &   5.44 & -  & - & -        \\
				\cite{ zhao2019cam} &   80.3 &   67.6 &   56.0 &  42.40 &   33.40 &     54.30 &    69.40 &                -   & -    & -  & - &  -       \\
				\cite{zheng2020syntax} & -    &     -&     -&  46.50 &   33.50 &     81.00 &    69.40 &            -       &   -  & -  & - &  -       \\
				\cite{wei2020exploiting} &   -  &    - &     -&  46.80 &   34.40 &     85.70 & -      &                  - &     -&  - & - & -        \\
				\bottomrule
			\end{tabular}
		}
		\caption{All results on Youtube2Text dataset}
		\label{apendice:ally2t}
	\end{table*}

	\begin{table*}[h]
		\resizebox{18cm}{!}
		{
			\begin{tabular}{lrrrrrrrrrrrr}
				\toprule
				Work &  BLEU-1 &  BLEU-2 &  BLEU-3 &  BLEU-4 &  METEOR &  CIDERr-D &  ROUGE-L &  Average-Recall (AR) &  SPICE &  FCE &  RE &  Self-BLEU \\
				\midrule
				\cite{fakoor2016memory} &   50.0 &   31.1 &   18.8 &   11.5 &    17.6 &      16.7 &      - &                  - &    - &  - & - &        - \\
				\cite{ li2017mam} &   50.6 &   31.7 &   21.3 &   12.7 &    19.1 &      18.3 &      - &                  - &    - &  - & - &        - \\
				\cite{ wu2018interpretable} &    - &    - &    - &   13.5 &    17.8 &      20.8 &      - &                  - &    - &  - & - &        - \\
				\cite{zhao2018video} &   50.7 &   31.3 &   19.7 &   13.3 &    19.0 &      18.0 &      - &                  - &    - &  - & - &        - \\
				\cite{ wang2018video} &   64.4 &   44.3 &   29.4 &   18.8 &    18.7 &      23.6 &     31.2 &                  - &    - &  - & - &        - \\
				\cite{hu2019hierarchical} &    - &    - &    - &   14.5 &    18.4 &      23.7 &      - &                  - &    - &  - & - &        - \\
				\cite{ zhao2019cam} &   51.3 &   32.1 &   21.7 &   12.9 &    19.7 &      18.8 &      - &                  - &    - &  - & - &        - \\
				\cite{wei2020exploiting} &    - &    - &    - &   12.7 &    17.2 &      21.6 &      - &                  - &    - &  - & - &        - \\
				\bottomrule
			\end{tabular}
		}
		\caption{All results on Charades dataset}
		\label{apendice:allcharades}
	\end{table*}

	\begin{table*}[h]
		\resizebox{18cm}{!}
		{
			\begin{tabular}{lrrrrrrrrrrrr}
				\toprule
				Work &  BLEU-1 &  BLEU-2 &  BLEU-3 &  BLEU-4 &METEOR &  CIDERr-D &  ROUGE-L &  Average-Recall (AR) &  SPICE &  FCE &  RE &  Self-BLEU \\
				\midrule
				\cite{yu2016video} &   60.8 &   49.6 &   38.5 &   30.5 &    28.7 &      1.62 &      - &                  - &    - &  - & - &        - \\
				\bottomrule
			\end{tabular}
		}
		\caption{All results on TACoS-MultiLevel dataset}
		\label{apendice:tacosmultinivel}
	\end{table*}

	\begin{table*}[h]
		\resizebox{18cm}{!}
		{
			\begin{tabular}{lrrrrrrrrrrrr}
				\toprule
				Work &  BLEU-1 &  BLEU-2 &  BLEU-3 &  BLEU-4 & METEOR &  CIDERr-D &  ROUGE-L &  Average-Recall (AR) &  SPICE &  FCE &  RE &  Self-BLEU \\
				\midrule
				\cite{yu2017end} &   13.5 &    4.4 &    1.7 &    0.8 &     7.1 &      10.0 &     15.9 &                  - &    - &  - & - &        - \\
				\cite{yu2017supervising} &   16.8 &    5.5 &    2.1 &    - &     7.2 &       9.3 &     15.6 &                  - &    - &  - & - &        - \\
				\cite{gao2019hierarchical} &    - &    - &    - &    0.7 &     5.6 &      10.4 &     14.6 &                  - &    - &  - & - &        - \\

				\bottomrule
			\end{tabular}
		}
		\caption{All results on LSMDC dataset}
		\label{apendice:LSMCD}
	\end{table*}

	\begin{table*}[h]
		\resizebox{18cm}{!}
		{
			\begin{tabular}{lrrrrrrrrrrrr}
				\toprule
				Work &  BLEU-1 &  BLEU-2 &  BLEU-3 &  BLEU-4 &  METEOR &  CIDERr-D &  ROUGE-L &  Average-Recall (AR) &  SPICE &  FCE &  RE &  Self-BLEU \\
				\midrule
				\cite{yu2017supervising} &   30.6 &   12.5 &    4.9 &    - &     8.4 &       8.4 &     22.9 &                  - &    - &  - & - &        - \\

				\bottomrule
			\end{tabular}
		}
		\caption{All results on VAS dataset}
		\label{apendice:vas}
	\end{table*}

	\begin{table*}[h]
		\resizebox{18cm}{!}
		{
			\begin{tabular}{lrrrrrrrrrrrr}
				\toprule
				Work &  BLEU-1 &  BLEU-2 &  BLEU-3 &  BLEU-4 & METEOR &  CIDERr-D &  ROUGE-L &  Average-Recall (AR) &  SPICE &  FCE &  RE &  Self-BLEU \\
				\midrule
				\cite{zhou2018end} &    - &    - &    - &   1.42 &   11.20 &       - &      - &                  - &    - &  - & - &        - \\
				\cite{yu2021accelerated} &    - &    - &    - &    - &    2.43 &      4.88 &      - &                  - &    - &  - & - &        - \\

				\bottomrule
			\end{tabular}
		}
		\caption{All results on YouCook2 dataset}
		\label{apendice:yc2}
	\end{table*}

	\begin{table*}[h]
		\resizebox{18cm}{!}
		{
			\begin{tabular}{lrrrrrrrrrrrr}
				\toprule
				Work &  BLEU-1 &  BLEU-2 &  BLEU-3 &  BLEU-4 & METEOR &  CIDERr-D &  ROUGE-L &  Average-Recall (AR) &  SPICE &  FCE &  RE &  Self-BLEU \\
				\midrule
				yu2018fine &  55.88 &  46.12 &  39.21 &  34.45 &   27.57 &      2.61 &     53.5 &                  - &   39.1 & 19.44 & - &        - \\

				\bottomrule
			\end{tabular}
		}
		\caption{All results on FSN dataset}
		\label{apendice:fsn}
	\end{table*}

	\begin{table*}[h]
		\resizebox{18cm}{!}
		{
			\begin{tabular}{lrrrrrrrrrrrr}
				\toprule
				Work &  BLEU-1 &  BLEU-2 &  BLEU-3 &  BLEU-4 & METEOR &  CIDERr-D &  ROUGE-L &  Average-Recall (AR) &  SPICE &  FCE &  RE &  Self-BLEU \\
				\midrule
				\cite{zhang2020object} &    - &    - &    - &   32.1 &    22.2 &      49.7 &     48.9 &                  - &    - &  - & - &        - \\
				\cite{chen2021motion} &    - &    - &    - &   34.2 &    23.5 &      57.6 &     50.3 &                  - &    - &  - & - &        - \\

				\bottomrule
			\end{tabular}
		}
		\caption{All results on VATEX dataset}
		\label{apendice:VATEX}
	\end{table*}

	\begin{table*}[h]
		\resizebox{18cm}{!}
		{
			\begin{tabular}{lrrrrrrrrrrrr}
				\toprule
				Work &  BLEU-1 &  BLEU-2 &  BLEU-3 &  BLEU-4 & METEOR &  CIDERr-D &  ROUGE-L &  Average-Recall (AR) &  SPICE &  FCE &  RE &  Self-BLEU \\
				\midrule
				\cite{qi2019sports} &    - &    - &    - &  31.76 &   26.07 &      2.91 &    51.62 &                  - &    - &  - & - &        - \\

				\bottomrule
			\end{tabular}
		}
		\caption{All results on SVCDV dataset}
		\label{apendice:SVCDV}
	\end{table*}

	\begin{table*}[h]
		\resizebox{18cm}{!}
		{
			\begin{tabular}{lrrrrrrrrrrrr}
				\toprule
				Work &  BLEU-1 &  BLEU-2 &  BLEU-3 &  BLEU-4 &  METEOR &  CIDERr-D &  ROUGE-L &  Average-Recall (AR) &  SPICE &  FCE &  RE &  Self-BLEU \\
				\midrule
				\cite{zhu2020understanding} &   49.2 &   35.5 &   27.2 &   21.2 &      21 &      54.3 &     47.5 &                  - &    - &  - & - &        - \\

				\bottomrule
			\end{tabular}
		}
		\caption{All results on Object-oriented captions dataset}
		\label{apendice:objocap}
	\end{table*}

	\begin{table*}[h]
		\resizebox{18cm}{!}
		{
			\begin{tabular}{lrrrrrrrrrrrr}
				\toprule
				Work &  BLEU-1 &  BLEU-2 &  BLEU-3 &  BLEU-4 & METEOR &  CIDERr-D &  ROUGE-L &  Average-Recall (AR) &  SPICE &  FCE &  RE &  Self-BLEU \\
				\midrule

				\cite{inacio2021osvidcap} &    - &    - &    - &  69.54 &   49.34 &    354.04 &    84.05 &                  - &    - &  - & - &        - \\

				\bottomrule
			\end{tabular}
		}
		\caption{All results on LIRIS dataset}
		\label{apendice:LIRIS}
	\end{table*}

	\begin{table*}[h]
		\resizebox{18cm}{!}
		{
			\begin{tabular}{lrrrrrrrrrrrr}
				\toprule
				Work &  BLEU-1 &  BLEU-2 &  BLEU-3 &  BLEU-4 &  METEOR &  CIDERr-D &  ROUGE-L &  Average-Recall (AR) &  SPICE &  FCE &  RE &  Self-BLEU \\
				\midrule

				\cite{dilawari2021natural} &    - &    - &    - &    - &    33.9 &       - &     0.72 &                  - &    - &  - & - &        - \\

				\bottomrule
			\end{tabular}
		}
		\caption{All results on TRECViD dataset}
		\label{apendice:TRECViD}
	\end{table*}

	\bibliographystyle{plain}

\end{document}